\pdfoutput=1
\documentclass[11pt]{article}
\usepackage[final]{acl}
\usepackage{times}
\usepackage{latexsym}
\usepackage[T1]{fontenc}
\usepackage[utf8]{inputenc}
\usepackage{microtype}
\usepackage{inconsolata}
\usepackage{graphicx}
\usepackage{mathtools} 
\usepackage{colortbl}
\usepackage{amsfonts}
\usepackage{graphicx} % For \resizebox
\usepackage[table]{xcolor} % For \rowcolor
\usepackage{stfloats}
\usepackage{float}
\usepackage{subcaption}
\usepackage{booktabs}

\title{SMEC:Rethinking Matryoshka Representation Learning for Retrieval Embedding Compression}

\author{Biao Zhang, Lixin Chen, Tong Liu \\
 Taobao \& Tmall Group of Alibaba \\
  Hangzhou, China \\
  \texttt{\{zb372670,tianyou.clx,yingmu\}@taobao.com} \\\And
  Bo Zheng \\
  Taobao \& Tmall Group of Alibaba \\
  Beijing, China \\
  \texttt{bozheng@alibaba-inc.com} \\}

\begin{document}
\maketitle

\begin{abstract}
% Large language models (LLMs) generate high-dimensional embeddings that capture rich semantic and syntactic information. However, due to the curse of dimensionality, high-dimensional embeddings exacerbate computational complexity and storage requirements, thereby hindering practical deployment. To resolve these bottlenecks, We propose Sequential Matryoshka Embedding Compression(SMEC), a novel sequential training structure to achieves significant dimensionality reduction without compromising performance, leading to marked improvements in computational efficiency and resource utilization. Also, we developed an adaptive dimension selection (ADS) mechanism, designed to preserve critical information during embedding compression by dynamically optimizing the retained feature subspaces. Additionally, overcome the scarcity of negative samples within a single batch, we present Selectable Cross-Batch Sample Mining (S-XBM), which dynamically retains historical feature embeddings through priority-based sampling, thereby enlarging the negative pool and boosting contrastive learning performance. Remarkably, extensive experiments across multi-modal datasets demonstrate that Sequential Matryoshka Embedding Compression achieves significant dimensionality reduction while maintaining performance.

Large language models (LLMs) generate high-dimensional embeddings that capture rich semantic and syntactic information. However, high-dimensional embeddings exacerbate computational complexity and storage requirements, thereby hindering practical deployment. To address these challenges, we propose a novel training framework named Sequential Matryoshka Embedding Compression (SMEC). This framework introduces the Sequential Matryoshka Representation Learning(SMRL) method to mitigate gradient variance during training, the Adaptive Dimension Selection (ADS) module to reduce information degradation during dimension pruning, and the Selectable Cross-batch Memory (S-XBM) module to enhance unsupervised learning between high- and low-dimensional embeddings. Experiments on image, text, and multimodal datasets demonstrate that SMEC achieves significant dimensionality reduction while maintaining performance. For instance, on the BEIR dataset, our approach improves the performance of compressed LLM2Vec embeddings (256 dimensions) by 1.1 points and 2.7 points compared to the Matryoshka-Adaptor and Search-Adaptor models, respectively.
\end{abstract}

\section{Introduction}
\begin{figure}[t!]
    \centering
    \includegraphics[width=\linewidth]{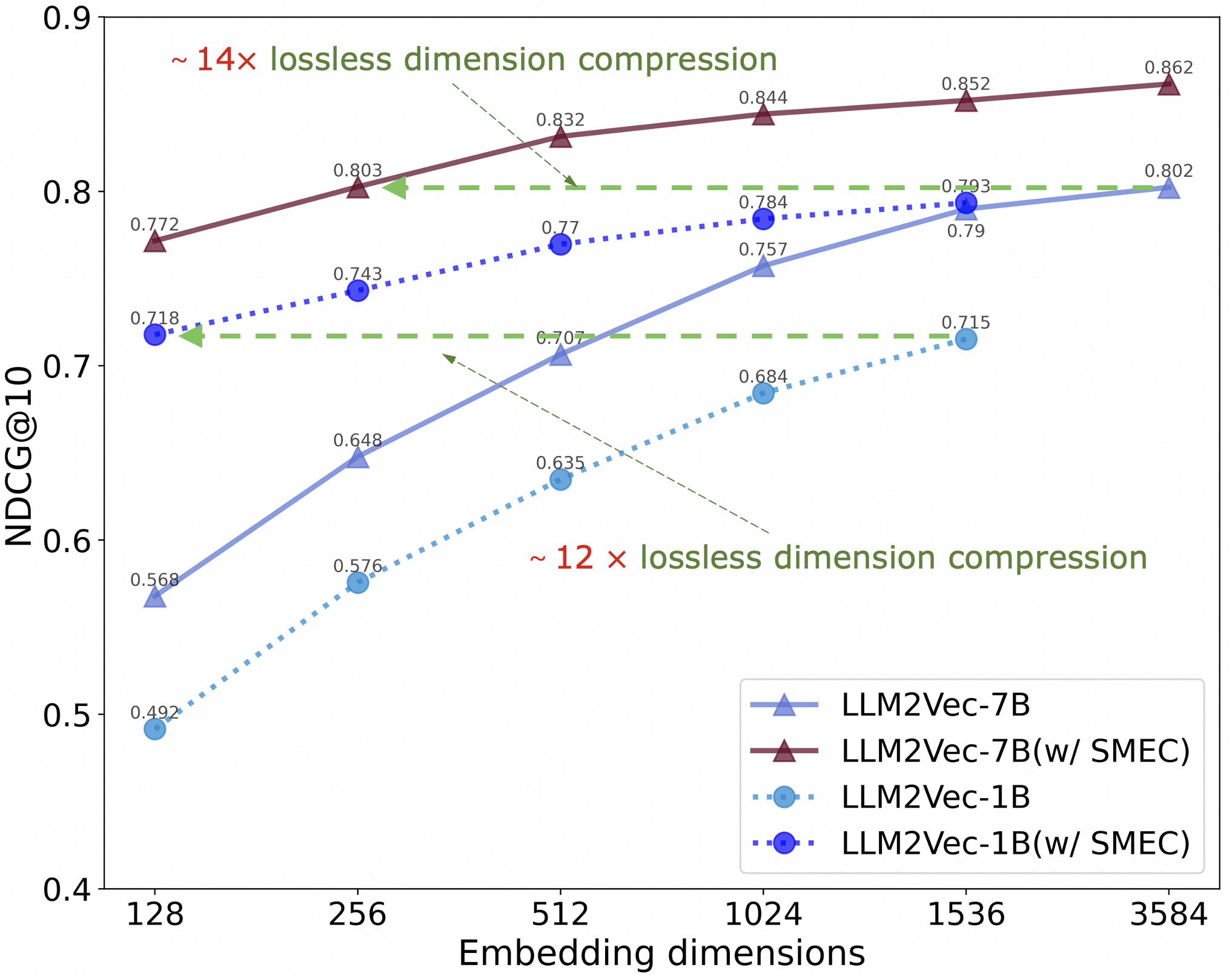}
    \caption{The effectiveness of the SMEC in dimensionality reduction. After customized training with the SMEC method on BEIR Quora dataset, the embeddings of LLM2Vec-7B (3584 dimensions) and LLM2Vec-1B (1536 dimensions) can achieve 14$\times$ and 12$\times$ lossless compression, respectively.}
    \label{fig:intr}
\end{figure}

\begin{figure*}[t!]
% {r}{0.5\textwidth}
\centering
\includegraphics[width=0.99\textwidth]{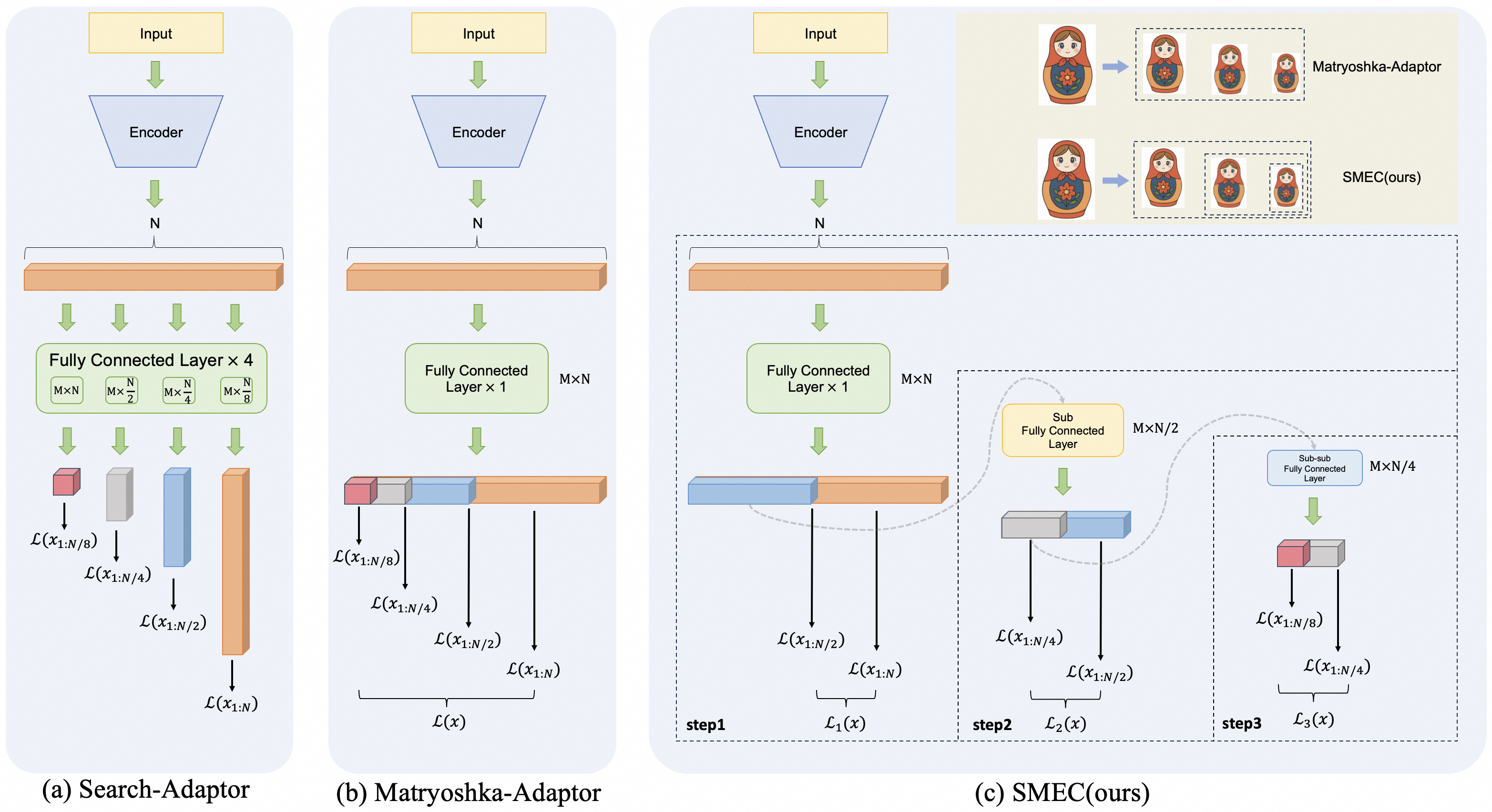}
\caption{Illustration of embedding compression architectures and our proposed approach. (a) presents the direct feature dimensionality reduction performed by the Search-Adaptor using FC layers. (b) illustrates the Matryoshka-Adaptor, which employs a shared set of FC layers to generate low-dimensional embeddings with multiple output dimensions. A Matryoshka-like hierarchical inclusion relationship exists between the high- and low-dimensional embeddings. (c) presents our proposed Sequential Matryoshka Embedding Compression (SMEC) framework, which adopts a sequential approach to progressively reduce high-dimensional embeddings to the target dimension. The animated diagram in the upper-right corner vividly highlights the distinction between Matryoshka-Adaptor and SMEC.
}
\label{fig:overview}
\end{figure*}

Large language models excel in diverse text tasks due to their ability to capture nuanced linguistic structures and contextual dependencies. For instance, GPT-4 achieves state-of-the-art performance on benchmarks like GLUE \cite{2018GLUE} and SuperGLUE \cite{wang2019superglue}, demonstrating its proficiency in tasks such as natural language inference (NLI), question answering (QA), and text classification. This success is attributed to their transformer-based architectures \cite{vaswani2017attention}, which enable parallel processing of sequential data and capture long-range dependencies through self-attention mechanisms. Similarly, Llama-3 \cite{grattafiori2024llama} and ChatGPT \cite{brown2020language} leverage similar principles to achieve comparable or superior performance in domain-specific and multi-lingual tasks.

% LLMs have also disrupted information retrieval (IR) by addressing limitations of traditional keyword-based systems. In dense retrieval frameworks like DPR \cite{karpukhin2020dense} and ColBERT \cite{khattab2020colbert}, LLM-derived embeddings outperform sparse term vectors in tasks such as document ranking and passage retrieval. For instance, Llama-3 achieves 85\% accuracy on the MS MARCO passage ranking task \cite{nguyen2016ms}, surpassing earlier models by capturing semantic nuances in queries and documents. Recent advances with LLM2Vec \cite{behnamghader2024llm2vec} further push the boundaries of dense retrieval. This framework employs a bidirectional attention mechanism of a pre-trained LLM and adopts the masked next-token prediction strategy for unsupervised representation training. Notably, LLM2Vec demonstrates statistically significant performance improvements across the full MTEB benchmark, with an absolute enhancement of 16.4\% observed for the Mistral-7B model compared to the baseline. Furthermore, LLMs demonstrate superior performance in cross-modal retrieval, linking text to images or videos \cite{radford2021learning}, and multilingual retrieval.

LLMs are increasingly integrated into commercial information retrieval (IR) systems, such as search engines (e.g., Google’s MUM) and recommendation platforms (e.g., Netflix’s content retrieval). Their ability to generate embeddings for long documents (e.g., books, research papers) and dynamic queries (e.g., conversational search) makes them indispensable for modern applications. For example, the BEIR benchmark \cite{2021BEIR} evaluates cross-domain retrieval performance, where LLMs outperform traditional BM25\cite{1994Some} and BERT-based models\cite{2019BERT} by leveraging contextual embeddings.

While LLMs’ high-dimensional embeddings enable sophisticated semantic modeling, their storage and computational costs hinder scalability. Embedding dimensions of LLMs typically range from 1,024 (e.g., GPT-3) to 4,096 (e.g., Llama-3), exacerbating storage overhead and computational inefficiency—especially in real-time systems requiring dynamic updates. 
Moreover, high-dimensional vectors degrade the performance of retrieval algorithms due to the curse of dimensionality \cite{1999When}. For example: exact nearest-neighbor search in high-dimensional spaces becomes computationally infeasible, necessitating approximate methods like FAISS \cite{2017Billion} or HNSW \cite{Yury2018Efficient}. Even with optimizations, query latency increases exponentially with dimensionality, limiting responsiveness in real-world applications. 

To address these challenges, Matryoshka Representation Learning (MRL) \cite{kusupati2022matryoshka} encodes multi-scale information into a single embedding, balancing task complexity and efficiency. It achieves strong results in large-scale classification and retrieval tasks and has inspired variants like Matryoshka-Adaptor \cite{yoon2024matryoshka}, which offers a scalable framework for transforming embeddings into structured representations with Matryoshka properties under both supervised and unsupervised settings. However, MRL's multi-scale parallel training strategy simultaneously limits its practical application in industry. When the retrieval system requires a new low-dimensional embedding, retraining from scratch is necessary to achieve effective dimensionality reduction.

% To address these challenges, Matryoshka Representation Learning (MRL) \cite{kusupati2022matryoshka} proposes a flexible approach that integrates multi-scale information into a single embedding vector. Its adaptive architecture enables simultaneous optimization of task-specific complexity and computational efficiency, which is validated through strong performance in large-scale adaptive classification and retrieval scenarios. It has since inspired a series of improved variants, such as Matryoshka-Adaptor \cite{yoon2024matryoshka}. The Matryoshka-Adaptor framework provides a scalable solution that enables the systematic transformation of arbitrary embedding spaces into structured representations with well-defined Matryoshka properties, operable under both unsupervised and supervised learning configurations. 

In this paper, we systematically analyze the limitations of MRL and its variants in embedding compression and propose three key enhancements: (1) a continued-training-friendly training framework named Sequential Matryoshka Representation Learning (SMRL); (2) an adaptive dimension selection (ADS) mechanism to minimize information degradation during dimension pruning; and (3) a Selectable Cross-batch Memory (S-XBM) strategy to enhance unsupervised learning between high- and low-dimensional embeddings.

% In this paper, we analyze the issues with using MRL for embedding compression and propose corresponding solutions. Our methodology makes three main contributions: First, we reform the Matryoshka training framework to mitigate gradient variance during training. Second, we propose an adaptive dimension selection (ADS) mechanism to minimize information degradation during feature compression. Third, we introduce selectable cross-batch sample mining (S-XBM) to construct a memory-augmented feature space through sequential embedding preservation. 

\section{Related Work}

\subsection{Matryoshka representation learning}
Matryoshka representation learning introduces a novel paradigm where embeddings are pretrained to inherently support progressive dimension truncation. This enables fine-grained control over the trade-off between computational latency (via reduced dimensionality) and accuracy (via retained semantic structure). Key innovations include the design of Matryoshka properties, such as hierarchical information encoding and intra-cluster compactness, which ensure that even truncated embeddings retain utility for downstream tasks.

In addition to representation learning, the concept of MRL have been applied to image generation, such as Matryoshka Diffusion Models (MDM) \cite{gu2023matryoshka}; multimodal content understanding, such as  $M^3$ \cite{cai2024matryoshka}; and Multimodal Large Language Model (MLLM), such as Matryoshka Query Transformer (MQT) \cite{hu2024matryoshka}.

% Beyond dimensionality reduction, MRL shares conceptual parallels with hashing-based methods \cite{1999Similarity}, which prioritize fast similarity search through compact codes. However, MRL differs by focusing on preserving semantic relationships across all truncation levels, whereas hashing often prioritizes binary codes for retrieval efficiency. 

% Recent advancements in multi-task learning and self-supervised learning have also contributed to the development of structured embeddings. For instance, frameworks like CLIP \cite{radford2021learning} leverage contrastive learning to align text and image embeddings, but they do not explicitly address the dimensionality truncation problem. The Matryoshka-Adaptor builds upon MRL by introducing hybrid unsupervised-supervised tuning strategies to further customize embeddings for dataset-specific characteristics, thereby surpassing the performance of embeddings derived purely from MRL pre-training.

\subsection{Embedding Compression}
Embedding compression aims to reduce the computational and memory footprint of neural network models or embeddings while preserving their utility for downstream tasks. This objective has driven research across multiple paradigms, each addressing different trade-offs between compression efficiency, performance retention, and adaptability.
Early approaches primarily focused on unsupervised techniques based on linear algebra, such as Principal Component Analysis (PCA) \cite{2016Principal}, Linear Discriminant Analysis (LDA) \cite{0Discriminant}, and Non-negative Matrix Factorization (NMF) \cite{lee2000algorithms}. Building upon these, autoencoders and their variants, such as Variational Autoencoders (VAEs) \cite{kingma2013auto}, have gradually emerged as powerful tools for nonlinear dimensionality reduction, capable of capturing complex data distributions. With the development of deep learning, methods such as Contrastive Predictive Coding (CPC) \cite{oord2018representation} and Momentum Contrast (MoCo) \cite{he2020momentum} are capable of learning robust and compact representations from unlabeled data.

% Early approaches focused on embedding-specific compression. Principal Component Analysis (PCA) \cite{2016Principal} remains a cornerstone for linear dimensionality reduction, leveraging orthogonal transformations to maximize variance retention. Supervised extensions like Linear Discriminant Analysis (LDA) \cite{0Discriminant} optimize class separability by maximizing inter-class variance, but they struggle with nonlinear decision boundaries and are sensitive to small-sample high-dimensional data. Autoencoders and their variants (e.g., Variational Autoencoders (VAEs) \cite{kingma2013auto}) emerged as powerful tools for nonlinear dimensionality reduction, capturing complex data distributions.However, existing self-supervised learning paradigms predominantly prioritize the fidelity of learned representations over explicit optimization for model compression efficiency.

Recently, customized methods such as Search-Adaptor \cite{yoon2023search} and Matryoshka-Adaptor \cite{yoon2024matryoshka} have emerged as a new trend in embedding compression. They achieve significant dimensionality reduction by adding only a small number of parameters to the original representation model and retraining it on specific data.

\section{Method}

\subsection{Rethinking MRL for embedding compression} 
\label{sec3.1}
MRL employs a nested-dimensional architecture to train models that learn hierarchical feature representations across multiple granularities. This allows adaptive deployment of models based on computational constraints. Specifically, MRL defines a series of models \(f_1, f_2, \ldots, f_M\) that share identical input and output spaces but progressively expand their hidden dimensions.

The term Matryoshka derives from the hierarchical parameter structure where the parameters of model \(f_m\) are nested within those of its successor \(f_{m+1}\). To illustrate, consider a FC layer within the largest model \(f_M\), which contains \(d_M\) neurons in its hidden layer. Correspondingly, the FC layer of \(f_m\) retains the first \(d_m\) neurons of this structure, with dimensions satisfying \(d_1 \leq d_2 \leq \dots \leq d_M\). MRL jointly trains these models using the following objective:  
\begin{equation}
\sum_{m=1}^M c_m \cdot \mathcal{L}(f_m(\mathbf{x}); y),
\end{equation}
where \(\mathcal{L}\) denotes the loss function, \(y\) represents the ground-truth label, and \(c_m\) are task-specific weighting coefficients. Notably, each training iteration requires forward and backward propagation for all \(M\) models, resulting in substantial computational overhead compared to training a single standalone model. Upon convergence, MRL enables flexible inference by selecting any intermediate dimension \(d_i \leq d_M\), thereby accommodating diverse computational constraints. 

Although the MRL method has partially mitigated the performance degradation of representations during dimensionality reduction, we contend that it still faces the following three unresolved issues:

\textbf{Gradient Fluctuation.} In large-scale vector retrieval systems, sample similarity is measured by the distance between their representation vectors. Consequently, the optimization of embedding models typically employs loss functions based on embedding similarity. In this condition, according to the derivation in Appendix \ref{appd:a}, the loss function $\mathcal{L}^d$ of MRL under dimension $d$ satisfies the following relationship with respect to the parameter $\mathbf{w}_i$ in the $i$-th dimension of the FC layer:

\begin{equation}
\frac{\partial \mathcal{L}^{d}}{\partial \mathbf{w}_i} \propto \frac{1}{\delta(d)^2}.
\end{equation} 
Here, $\delta(d)$ is a complex function that is positively correlated with the dimension $d$. This equation provides a mathematical foundation for analyzing gradient fluctuations in multi-dimensional joint optimization architecture. It indicates that during the MRL training process, loss functions from various dimensions result in gradients of varying magnitudes on the same model parameter, thereby increasing gradient variance. In Section \ref{sec5.2}, we empirically demonstrated that the conclusion above is applicable to different loss functions. We propose a solution to resolve the aforementioned problem in Section \ref{sec3.2}.

\textbf{Information Degradation.} Neural network parameters exhibit heterogeneous contributions to model performance, as demonstrated by the non-uniform distribution of their gradients and feature importance metrics \cite{frankle2018lottery}. The MRL method employs a dimension truncation strategy (e.g., \(D \rightarrow D/2 \rightarrow D/4 \ldots\)) to prune parameters and reduce feature dimensions by retaining partial parameters. However, this approach fails to adequately preserve critical parameters because it relies on a rigid, static truncation rule.
% rather than dynamically identifying parameter importance during training
Although MRL employs joint training of high- and low-dimensional vectors to redistribute information between truncated and retained parameters, this process is unavoidably accompanied by information degradation. Specifically, discarded parameters may contain essential information, such as unique feature mappings or high-order dependencies, that cannot be effectively recovered by the remaining ones. Empirical evidence, such as accuracy degradation and increased generalization gaps, demonstrates that such loss leads to suboptimal model performance and slower convergence \cite{li2023losparse}. 
% Notably, the static truncation strategy in MRL fails to adaptively prioritize parameters critical for capturing discriminative features and complex relationships. Although post-truncation training can redistribute some information, it is insufficient to fully recover the lost representational capacity. 
In summary, while MRL enables hierarchical dimensionality reduction, its inability to selectively retain critical parameters and the inherent information degradation during post-truncation training ultimately undermine its effectiveness in maintaining model performance. In Section \ref{sec3.3}, we propose a more effective dimension pruning method.

\begin{figure}[!h]
% {r}{0.5\textwidth}
\centering
\includegraphics[width=1\linewidth]{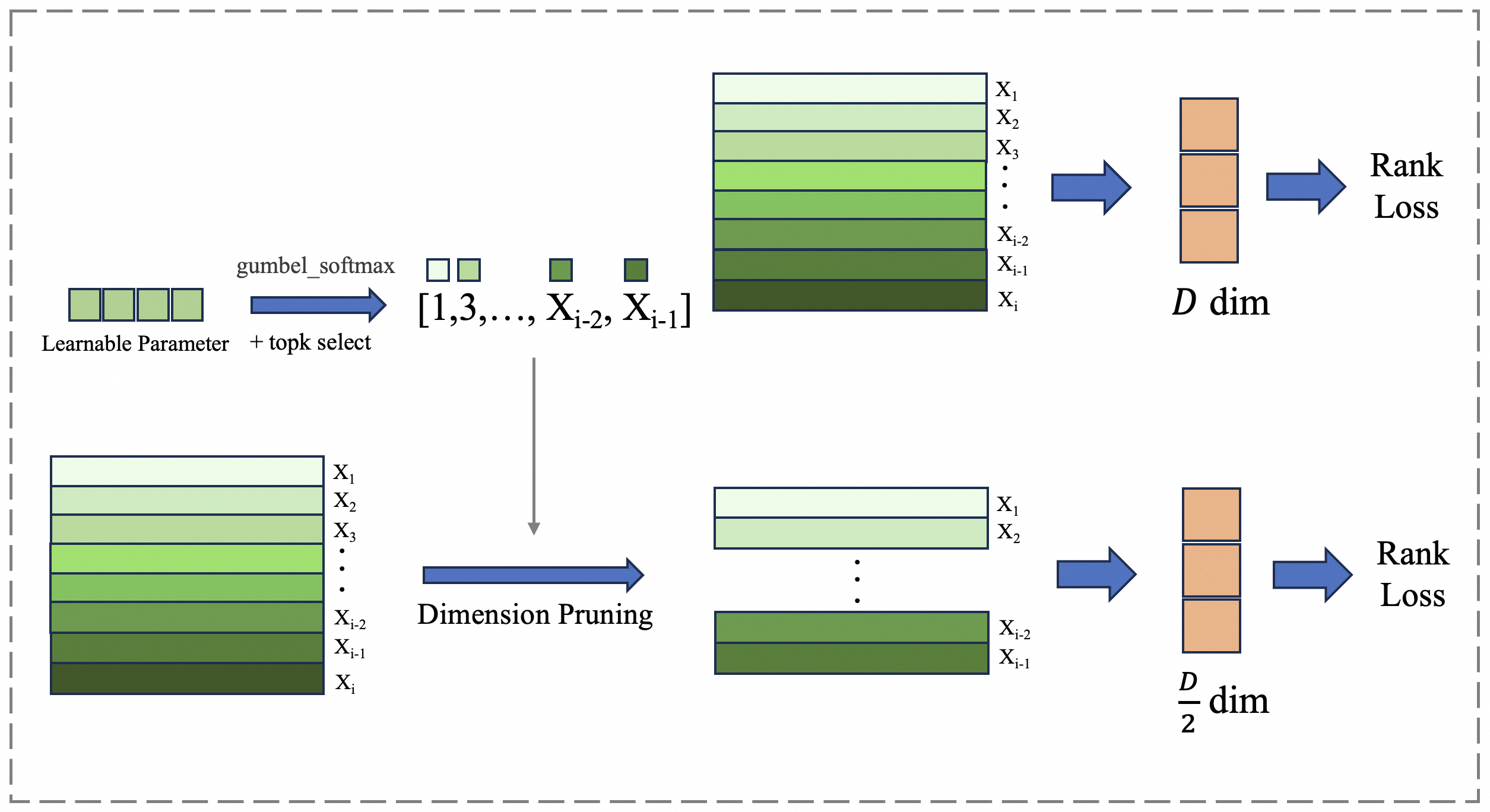}
\caption{The ADS module introduces a set of learnable parameters to dynamically select dimensions based on their importance during the dimensionality reduction process.
}
\label{fig:ads}
\end{figure}

\textbf{Sample Selection.} The MRL framework employs supervised learning to jointly train high-dimensional ($D$) and low-dimensional ($D'$) features. However, the number of available samples is limited by manual annotation. Matryoshka-Adaptor introduces in-batch sample mining strategies to expand the training sample scale, thereby addressing the inherent limitation. Specifically, it generates cross-sample pairs via the cartesian product of batch samples:  

\begin{equation}
\label{eq15}
\mathcal{P} = \{(x_i, x_j) \mid x_i, x_j \in \text{Batch},\ i \neq j\} .
\end{equation}
This approach creates $B(B-1)$ pairs per batch (where $B$ denotes the batch size), enabling cross-sample comparisons within large batches. However, this indiscriminate pairing introduces noise from non-representative or irrelevant sample pairs.  

In light of this limitation, the method employs Top-$k$ similarity-based selection:  
\begin{equation}
\label{eq11}
\begin{split}
\mathcal{P}_{\text{top-}k} &= \text{Top}_k\left(\text{similarity}(x_i, x_j)\right), \\
&\quad \forall\ (x_i, x_j) \in \mathcal{P}.
\end{split}
\end{equation}
Here, only the top-$k$ most similar pairs are retained for training, reducing computational overhead while focusing on informative interactions. Despite this improvement, the diversity of effective samples remains fundamentally constrained by the original batch size $B$. In Section \ref{sec3.4}, we develop a strategy that empowers the model to mine global sample beyond the current batch.
% Even with top-$k$ selection, the maximum number of usable pairs is limited to $O(B^2)$, which restricts the framework's ability to capture global distributional patterns beyond the batch-level context.  

\subsection{Sequential Matryoshka Representation Learning}
\label{sec3.2}
Applying the conclusions from Section \ref{sec3.1} to the MRL training process, and take the parallel dimensionality reduction process \([D, D/2, D/4]\) as an example. The ratio of the average gradients for parameters 
$\mathbf{w}_i(i \in [0,D/4])$ and $\mathbf{w}_j(j \in [D/4,D/2])$ is as follows:

\begin{align}
\label{eq12}
\overline{\text{grad}_i} 
&: \overline{\text{grad}_j} = \left( \frac{\partial \mathcal{L}^{D}}{\partial \mathbf{w}_i}  
+ \frac{\partial \mathcal{L}^{D/2}}{\partial \mathbf{w}_i} 
+ \frac{\partial \mathcal{L}^{D/4}}{\partial \mathbf{w}_i} 
\right) \notag \\
&: \left( \frac{\partial \mathcal{L}^{D}}{\partial \mathbf{w}_i}  
+ \frac{\partial \mathcal{L}^{D/2}}{\partial \mathbf{w}_i} \right) 
\approx 1 + \frac{\delta(D/2)^2}{\delta(D/4)^2} .
\end{align}

As shown in Equation \ref{eq12}, the average gradient magnitude of parameter $\mathbf{w}_i$ can be approximated as $1 + \frac{\delta(D/2)^2}{\delta(D/4)^2}$ times that of parameter $\mathbf{w}_j$, primarily due to the influence of the lower-dimensional loss function $\mathcal{L}^{D/4}$. To resolve this issue, we propose Sequential Matryoshka Representation Learning (SMRL), which substitutes the original parallel compression of embeddings with a sequential approach, as illustrated in the Figure \ref{fig:overview}. Assuming a dimensionality reduction trajectory of \([D, D/2, D/4, \dots, D/2^n]\). In each iteration, only the immediate transition (e.g., $D/2^{n-1} \rightarrow D/2^n$) is trained, avoiding the inclusion of lower-dimensional losses that amplify gradients for low-dimensional parameters. By eliminating the above factor, the gradients of $\mathbf{w}_i(i \in [0,D/2^n])$  follow a consistent distribution with reduced variance, improving convergence speed and performance. Once the loss converges in the current iteration, the dimensionality reduction \(D/2^{n-1} \rightarrow D/2^n\) is complete, and the process proceeds to the next stage \(D/2^n \rightarrow D/2^{n+1}\), repeating the procedure until the target dimension is reached. Additionally, after convergence in one iteration, the optimal parameters for the current dimension are fixed to prevent subsequent reductions from degrading their performance. Notably, compared to MRL, the SMRL framework is more amenable to continued training. In scenarios where low-dimensional retrieval embeddings (e.g., D/8) or intermediate embeddings (e.g., D/3) are required, these can be obtained through further dimensionality reduction training based on the already preserved D/4 or D/2 parameters, eliminating the need for retraining from scratch as is typically required in MRL.

\subsection{Adaptive Dimension Selection Module}

\label{sec3.3}
Since directly truncating dimensions to obtain low-dimensional representations in MRL inevitably leads to information degradation, we propose the Adaptive Dimension Selection (ADS) module to dynamically identify important dimensions during training. As illustrated in Figure \ref{fig:ads}, we introduce a set of learnable parameters that represent the importance of different dimensions in the original representation $\mathbf{Z} (\text{dim}=D)$, and use these parameters to perform dimensional sampling, obtaining a reduced-dimension representation $\mathbf{Z}^{\prime} (\text{dim}=D/2)$. Since the sampling operation is non-differentiable, during the training phase, we utilize the Gumbel-Softmax \cite{jang2016categorical} to approximate the importance of different dimensions. This is achieved by adding Gumbel-distributed noise \( G \sim \text{Gumbel}(0, 1) \) to the logits parameters $\hat{\mathbf{z}}$ for each dimension, followed by applying the softmax function to the perturbed logits to approximate the one-hot vector representing dimension selection. Mathematically, this can be expressed as:  
\begin{equation} 
\label{eq13}
\mathbf{z} = \text{softmax}_\tau(\hat{\mathbf{z}} + G).
\end{equation} 
Importantly, the Gumbel approximation allows the softmax scores of dimension importance to be interpreted as the probability of selecting each dimension, rather than enforcing a deterministic selection of the top-\(k\) dimensions. This achieves a fully differentiable reparameterization, transforming the selection of embedding dimensions into an optimizable process.

\begin{figure}[!t]
% {r}{0.5\textwidth}
\centering
\includegraphics[width=1\linewidth]{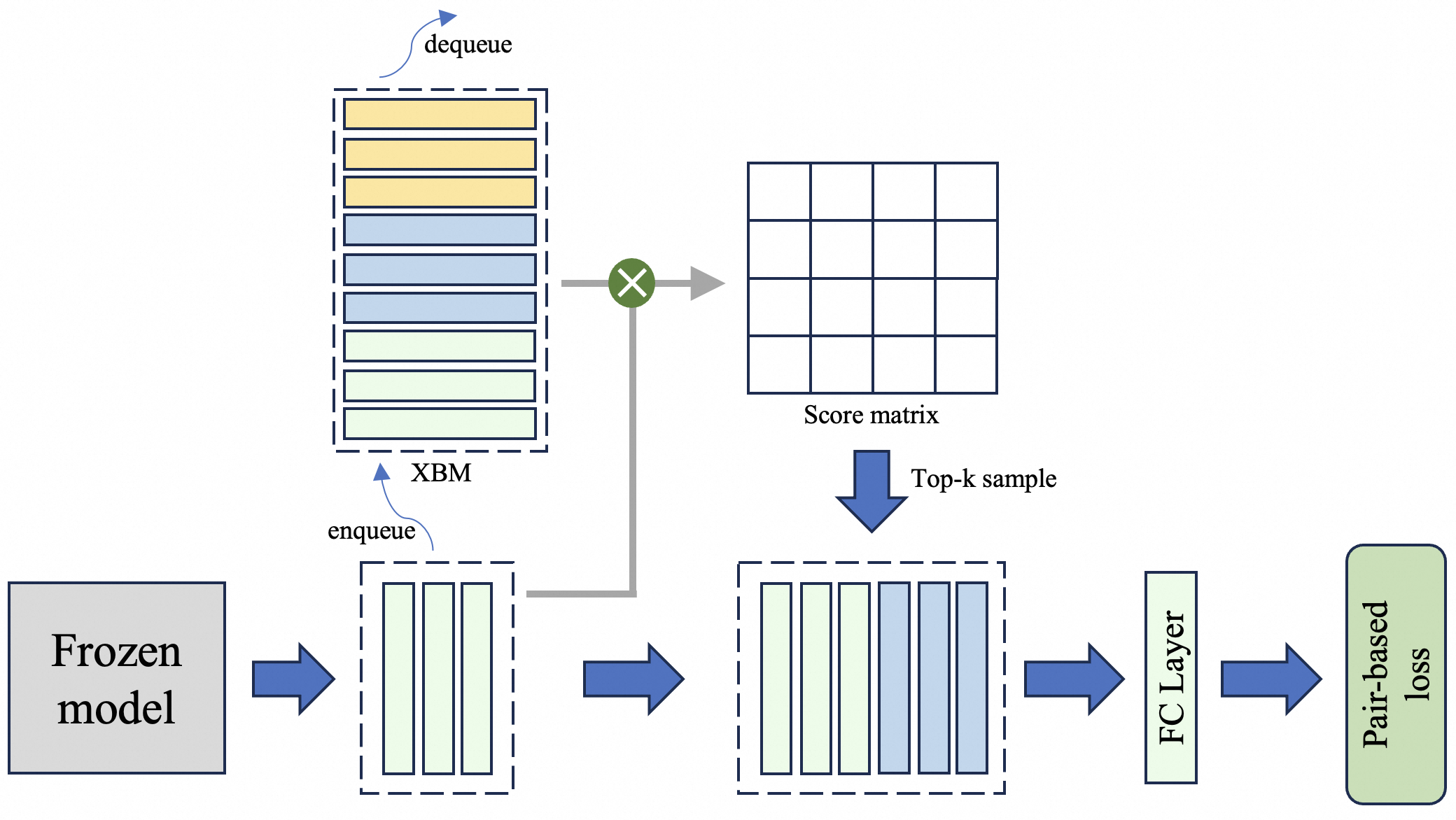}
\caption{S-XBM maintains a queue during training to store historical features across batches. Rather than incorporating all stored features into the current batch, it selectively leverages hard samples that exhibit high similarity to the current batch samples.
}
\label{fig:sxbm}
\end{figure}

\subsection{Selectable Cross-Batch Memory}
\label{sec3.4}
A natural teacher-student relationship inherently exists between the original embedding and its reduced-dimensional counterpart, making it feasible to improve the compressed embedding through unsupervised learning \cite{yoon2024matryoshka}. However, as discussed in Section \ref{sec3.1}, performing this process within a single batch suffers from sample noise and insufficient diversity. As illustrated in Figure \ref{fig:sxbm}, we propose the Selectable Cross-Batch Memory (S-XBM) module, which constructs a first-in-first-out (FIFO) queue during training to store original embeddings across batches, with the aim of addressing this limitation. Unlike the original XBM \cite{wang2020cross}, we introduce two task-specific improvements: (1) retrieving only the top‑$k$ most similar samples from the memory bank to construct new batches, and (2) deferring the trainable FC layer and only storing features generated by the frozen backbone, thereby avoiding feature drift.
The unsupervised loss between original embedding $emb$ and low-dimensional embedding $emb[:d]$ is as follows:
\begin{align}
{\mathcal{L}_{un-sup}} &= \sum_i \sum_{j \in \mathcal{N}_K(i)} 
    \left| \text{Sim}(emb_i, emb_j) \right. \notag \\
    &\quad \left. - \text{Sim}(emb_i[:d], emb_j[:d]) \right|
\end{align}
where $\mathcal{N}_K(i)$ denotes the set of the top $k$ most similar embeddings to $emb_i$ within the S-XBM module.

\begin{figure*}[t!]
\centering
\begin{subfigure}[t]{0.48\textwidth}
    \centering
    \includegraphics[width=\linewidth]{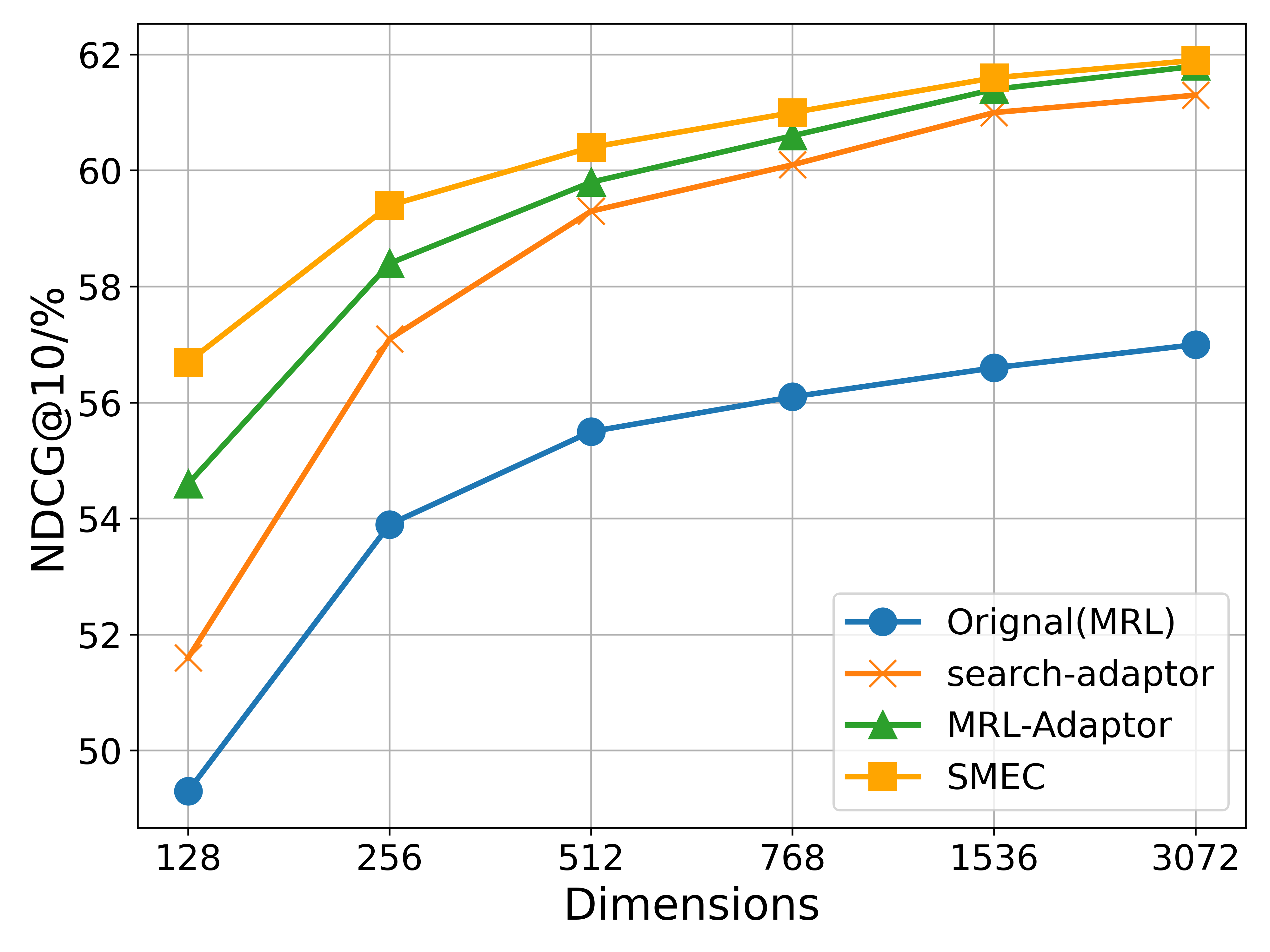}
    \caption{OpenAI text embeddings}
\end{subfigure}
\hfill
\begin{subfigure}[t]{0.48\textwidth}
    \centering
    \includegraphics[width=\linewidth]{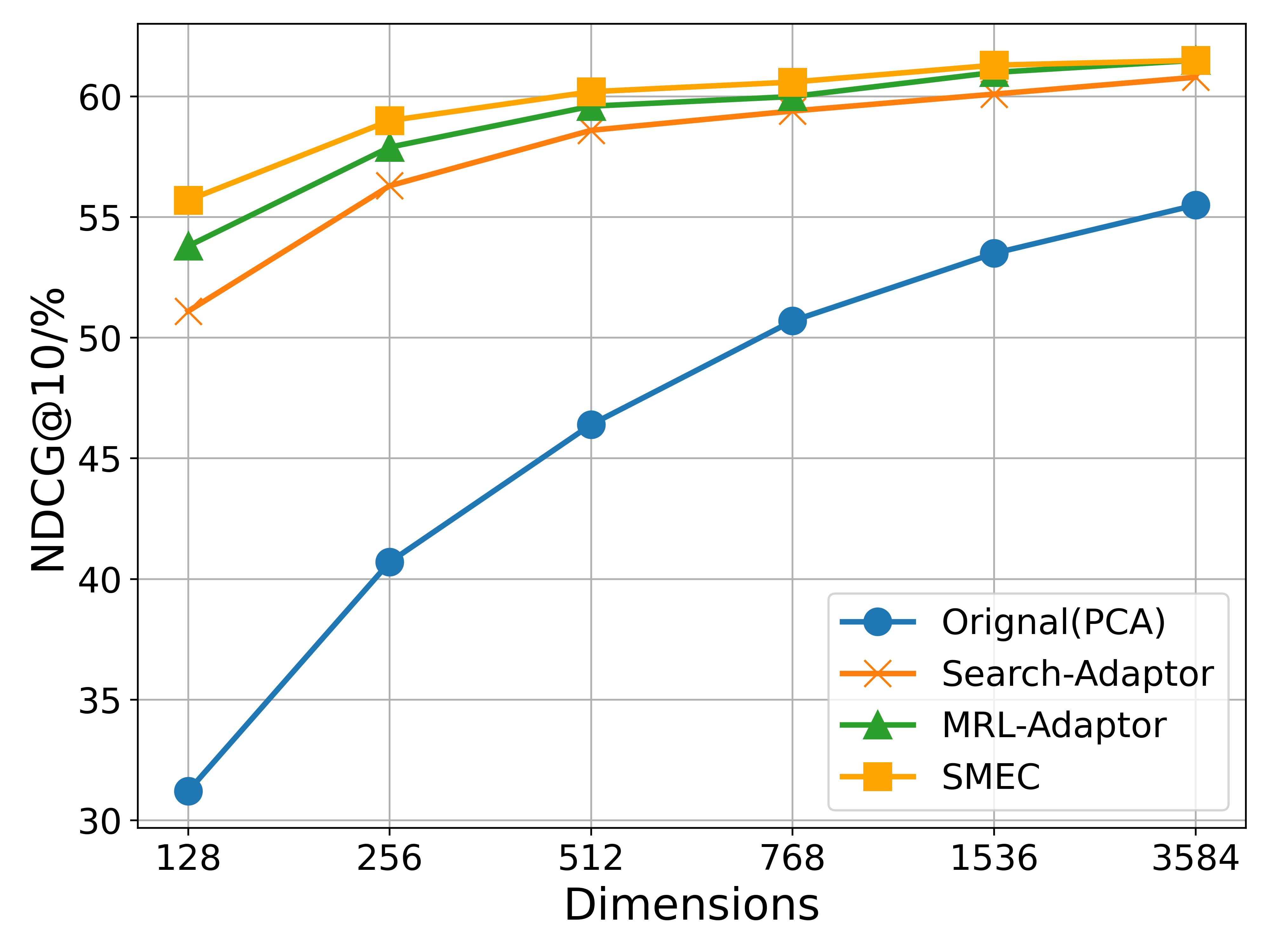}
    \caption{LLM2Vec}
\end{subfigure}
\caption{Experimental results on the BEIR dataset comparing two models: OpenAI’s text-embedding-3-large (with 3072 dimensions) and LLM2Vec (with 3548 dimensions), the latter built upon the Qwen2-7B model. OpenAI text embeddings inherently contain multi-scale representations (enabled by MRL during pretraining), while LLM2Vec obtains its orignal low-dimensional representations via PCA.}
\label{fig:exp2_1}
\end{figure*}

\section{Experiments}
In this section, we compare our approach with state-of-the-art methods in the field of embedding dimensionality reduction. 
\subsection{Dataset Description}
We evaluate the model's retrieval performance across diverse datasets: BEIR \cite{2021BEIR} (text retrieval), Products-10K \cite{2020Products} (image retrieval), and Fashion-200K \cite{2017Automatic} (cross-modal retrieval). BEIR is a comprehensive text retrieval benchmark consisting of 13 selected datasets from diverse domains. Products-10K contains approximately 10,000 products with over 150,000 images for large-scale product image retrieval. Fashion-200K includes over 200,000 fashion items with paired image-text data for cross-modal tasks.

% To evaluate the model's retrieval performance across diverse scenarios, we selected three benchmark datasets: BEIR (text retrieval), Products-10K \cite{2020Products}(image retrieval), and Fashion-200K \cite{2017Automatic}(cross-modal retrieval). 

% \textbf{BEIR} (Benchmarking IR) is a heterogeneous collection of benchmark datasets specifically formulated to assess the generalization performance of text-based information retrieval models. The benchmark integrates 18 publicly accessible datasets originating from disparate application domains and task formulations. Following Matryoshka-Adaptor, we curated a subset comprising 13 representative datasets; the statistics of these subsets are systematically documented in the Appendix.

% \textbf{Products-10K} is a benchmark dataset focused on large-scale product image retrieval, primarily used to evaluate the image search performance of models in complex e-commerce scenarios. It includes approximately 10,000 products with a total of over 100,000 images.

% \textbf{Fashion-200K} is a large-scale multimodal dataset in the fashion domain, specifically designed for tasks such as cross-modal retrieval (e.g., mutual image-text retrieval), product recommendation, and visual semantic understanding. The dataset is categorized into five subsets based on item categories, encompassing approximately 200,000 clothing product images paired with their corresponding textual descriptions.

\subsection{Implementation Details}

We use state-of-the-art models to extract the original embeddings for different datasets. Specifically, the BEIR dataset employs OpenAI text embeddings \cite{openai-text-embedding} and LLM2Vec \cite{behnamghader2024llm2vec} for text representation; the Products-10K dataset utilizes LLM2CLIP \cite{huang2024llm2clippowerfullanguagemodel} to obtain cross-modal embeddings; and the Fashion-200K dataset extracts image embeddings using the ViT-H\cite{dosovitskiy2020image} model. All dimensionality reduction methods are performed based on these original representations. To align with other methods, SMEC also adopts rank loss \cite{yoon2023search} as the supervised loss function, which is defined as follows:
\begin{align}
 \mathcal{L}_{rank} &= \sum_i\sum_j\sum_k\sum_m I(y_{ij} > y_{ik}) (y_{ij} - y_{ik}) \nonumber \\
    &\log(1+\exp(s_{ik}[:m] - s_{ij}[:m])),
\end{align}
where $I(y_{ij} > y_{ik})$ is an indicator function that is equal to 1 if $y_{ij}$ > $y_{ik}$ and 0 otherwise. $s_{ij}[:m]$ represents the cosine similarity between the query embedding $emb_i[:m]$ and the corpus embedding $emb_j[:m]$.
The total loss function is:
\begin{align}
  \mathcal{L}_{total} = \mathcal{L}_{rank} + \alpha\mathcal{L}_{un-sup} ,
\end{align}
with $\alpha$ being hyper-parameters with fixed values as $\alpha=1.0$. As SMEC involves multi-stage training, the training epochs of other methods are aligned with the total number of epochs costed by SMEC, and their best performance is reported.

% We utilize state-of-the-art models to extract original embeddings for different datasets. Specifically, the BEIR dataset employs Google Gecko text embeddings for textual representation. The Fashion-200K dataset uses Google Multimodal Embeddings to obtain cross-modal representations, while for the Products-10K dataset, image embeddings are extracted using the ViT-H model.All dimensionality reduction methods are performed based on the original representation, and all use rank loss as the loss function.
% In the inference stage, we first obtain the original embeddings using the aforementioned models. Then, different dimensionality reduction methods are applied to compress the embeddings. The retrieval performance of the compressed embeddings is evaluated using the Normalized Discounted Cumulative Gain (NDCG) \cite{jarvelin2002cumulated} metric.

\subsection{Results}
In this subsection, the results on the BEIR, Fashion-200K, and Products-10K datasets are given. Retrieval performance is evaluated using the normalized discounted cumulative gain at rank 10 (nDCG@10)\cite{jarvelin2002cumulated} metric.

\textbf{BEIR.} As shown in Figure \ref{fig:exp2_1}, we compare the performance of SMEC and other state-of-the-art methods on two types of models: the API-based OpenAI text embedding and the open-source LLM2vec, across various compressed dimensions. Significantly, SMEC exhibits the strongest performance retention, particularly at lower compression ratios. For example, when compressed to 128 dimensions, SMEC improves the performance of the OpenAI and LLM2vec models by 1.9 and 1.1 points respectively, compared to the best-performing Matryoshka-Adaptor.

\textbf{Products-10K.} Images naturally contain denser features than text \cite{o2020unsupervised}. As shown in Figure \ref{fig:img} of Appendix \ref{appd:b}, SMEC surpasses other dimensionality reduction methods in image retrieval tasks, highlighting the effectiveness of the ADS module in mitigating information degradation during dimension pruning. 

\textbf{Fashion-200K.} 
% In contrast to unimodal datasets, Fashion-200K is a multimodal dataset in which the queries and documents belong to different modalities — for instance, image-to-text or text-to-image retrieval. As shown in the figure, we compare the retrieval performance of various methods in both scenarios. The results indicate that SMEC achieves superior performance across both retrieval directions, further validating its robustness in multimodal contexts.
Unlike unimodal datasets, Fashion-200K involves cross-modal queries and documents, such as image-to-text and text-to-image retrieval. As illustrated in the Figure \ref{fig:t2i} and \ref{fig:i2t} of Appendix \ref{appd:b}, SMEC achieves superior performance in both directions, demonstrating strong robustness in multimodal scenarios.

\begin{figure*}[t!]
\centering
\begin{subfigure}[t]{0.31\textwidth}
    \centering
    \includegraphics[width=\linewidth]{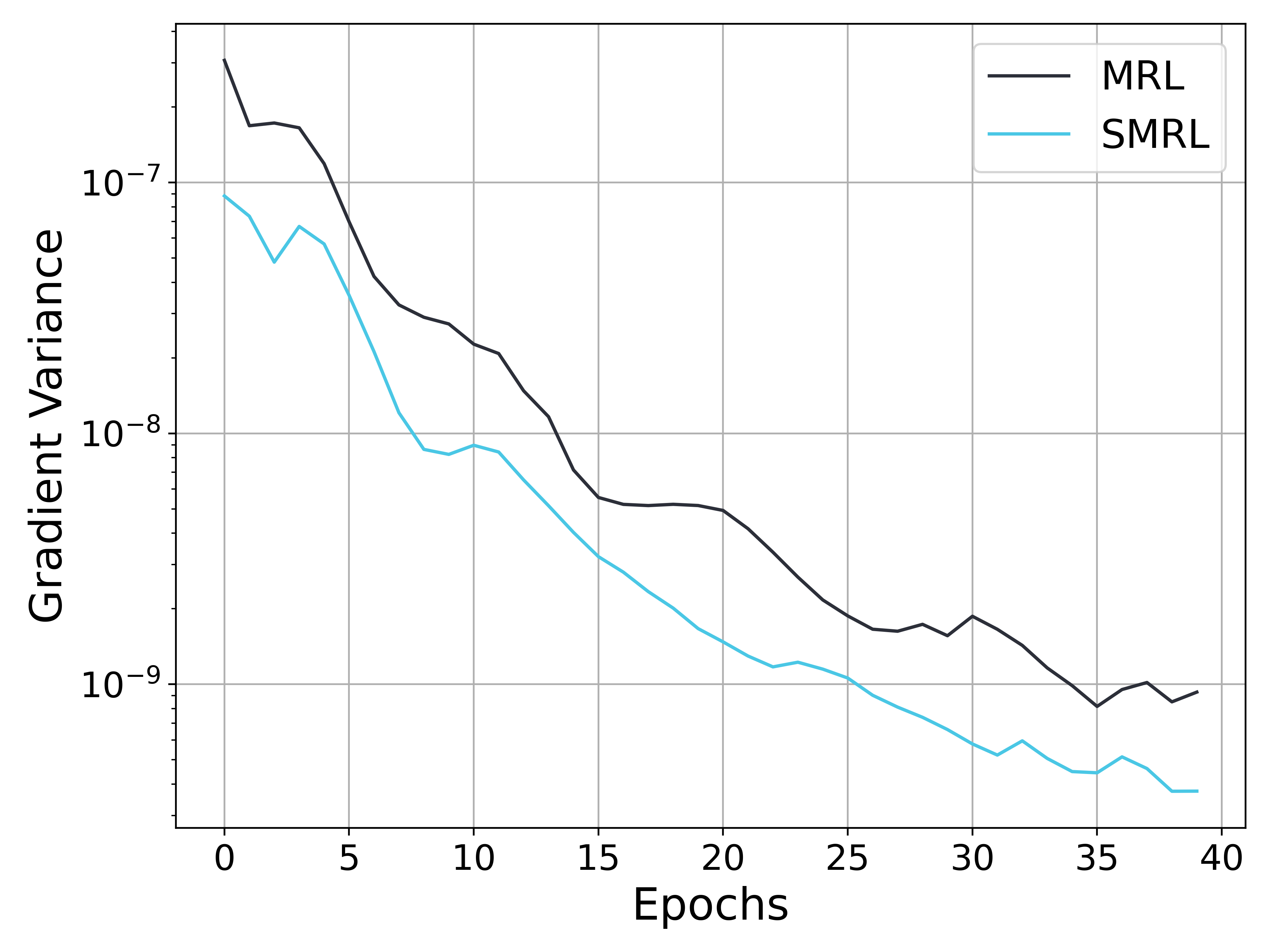}
    \caption{Gradient Variance}
    \label{fig:var}
\end{subfigure}
\hfill
\begin{subfigure}[t]{0.31\textwidth}
    \centering
    \includegraphics[width=\linewidth]{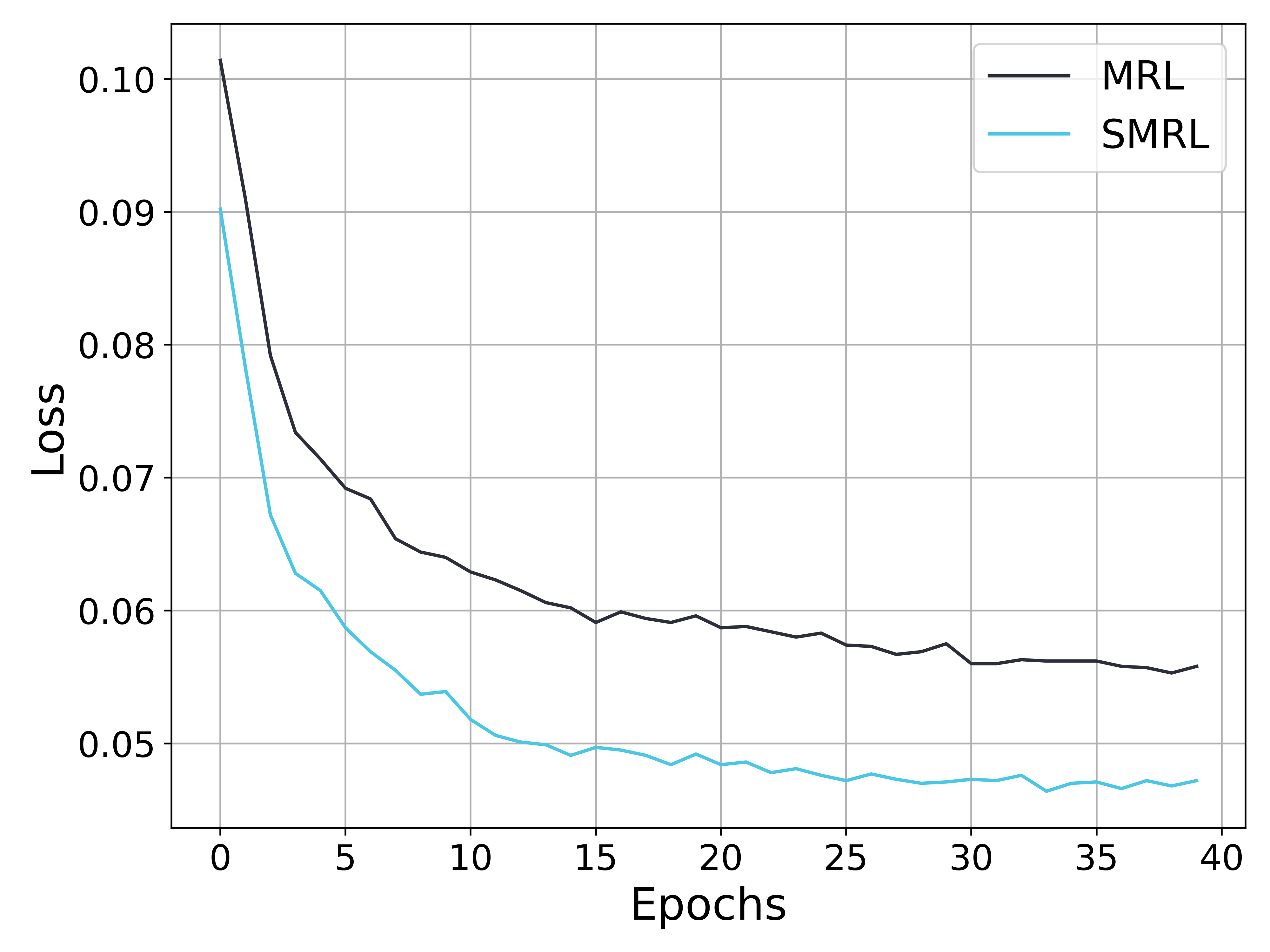}
    \caption{Validation loss}
    \label{fig:loss}
\end{subfigure}
\hfill
\begin{subfigure}[t]{0.31\textwidth}
    \centering
    \includegraphics[width=\linewidth]{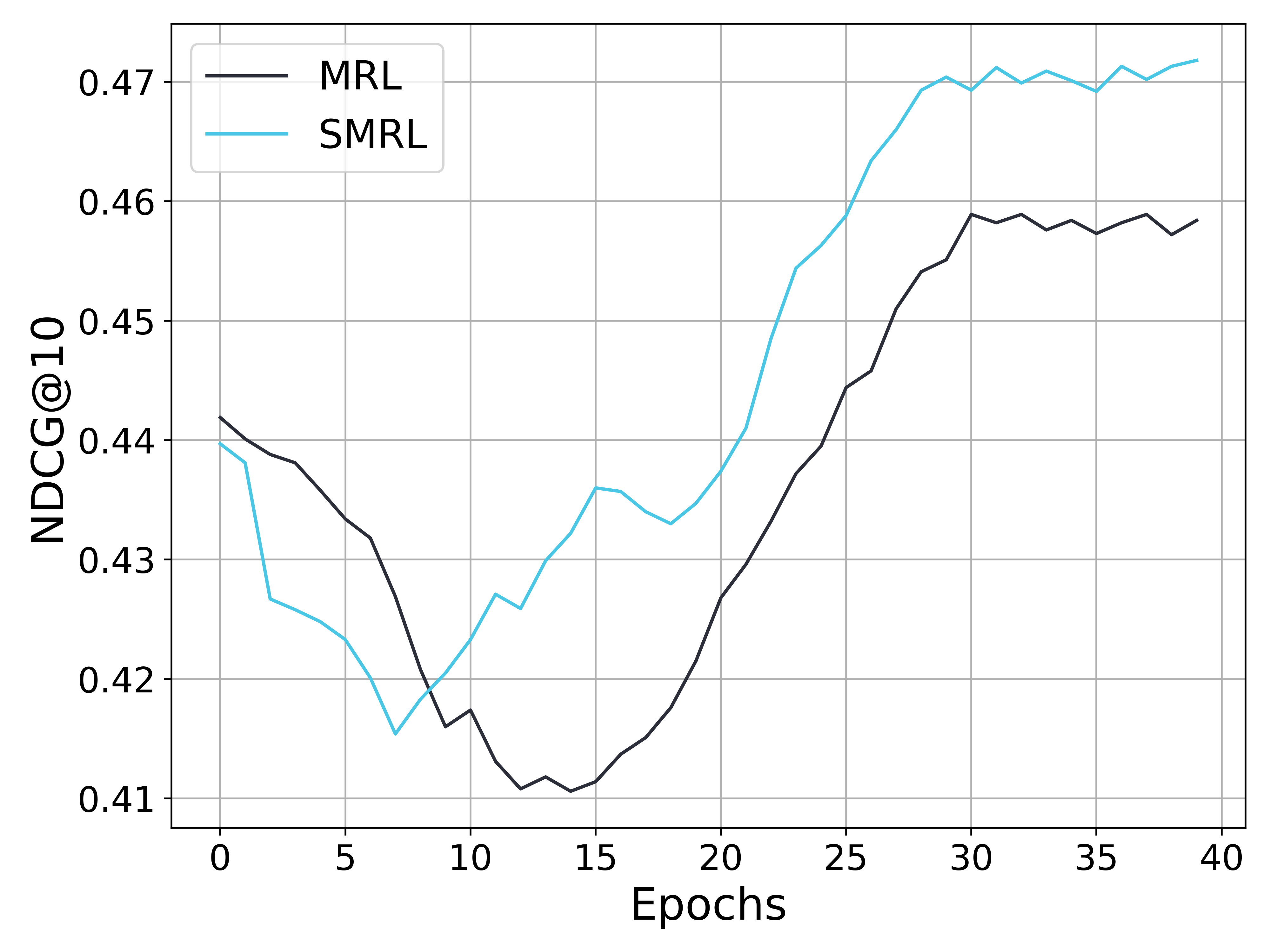}
    \caption{Retrieval performance}
    \label{fig:ndcg}
\end{subfigure}
\caption{Analysis of metrics during the training process. (a) shows the gradient variance curve (with the vertical axis in a logarithmic scale), (b) presents the loss curve on the validation set, and (c) illustrates the performance variations on the test set. As training progresses, the gradient variances of both MRL and SMRL decrease; however, the gradient variance of MRL remains several times higher than that of SMRL. Consequently, the loss curve of SMRL converges more quickly to a lower value, and the compressed embedding demonstrates better retrieval performance.}
\label{fig:var_loss_ndcg}
\end{figure*}

\section{Discussions}

\subsection{The influence of gradient variance}
\label{sec5.1}
To validate the impact of gradient variance on convergence speed and model performance (as discussed in Section \ref{sec3.2}), we conducted comparative experiments between SMRL and MRL using the MiniLM model on the BEIR dataset. As shown in Figure \ref{fig:var}, MRL consistently exhibits significantly higher gradient variance than SMRL throughout training. Consequently, the training loss of MRL continues to decline beyond the 20th epoch, whereas SMRL’s loss starts to converge at the 15th epoch. A similar trend is observed in subfigure \ref{fig:ndcg}, where SMRL enters the improvement phase earlier and converges to superior performance.

% To validate the impact of gradient variance on convergence dynamics and model performance (as discussed in Section \ref{sec3.2}), we employed the MiniLM model to conduct comparative experiments on the BEIR dataset using both SMRL and MRL (with target dimension 96). During training, we systematically analyzed gradient variance across model parameters spanning dimensions \({D \in [0,192]}\), alongside tracking training loss curves (as illustrated in the figure below). Notably, the Method 2 exhibited a markedly higher gradient magnitude compared to the Method 1 counterpart throughout the training process. While both methods demonstrated decreasing gradient variance as training progressed, the Method 2 maintained significantly larger variance (by several orders of magnitude) during the final stabilization phase. Corresponding to these observations, the loss trajectory reveals that the  Method 2 continued to exhibit descending trends beyond 20 epochs, whereas the  Method 1 entered a fluctuating convergence state by the 15th epoch. Importantly, the  Method 1 achieved a lower final loss value than the  Method 2, despite its earlier convergence.

\begin{figure}[t!]
\centering
\begin{subfigure}[t]{0.48\textwidth}
    \centering
    \includegraphics[width=\linewidth]{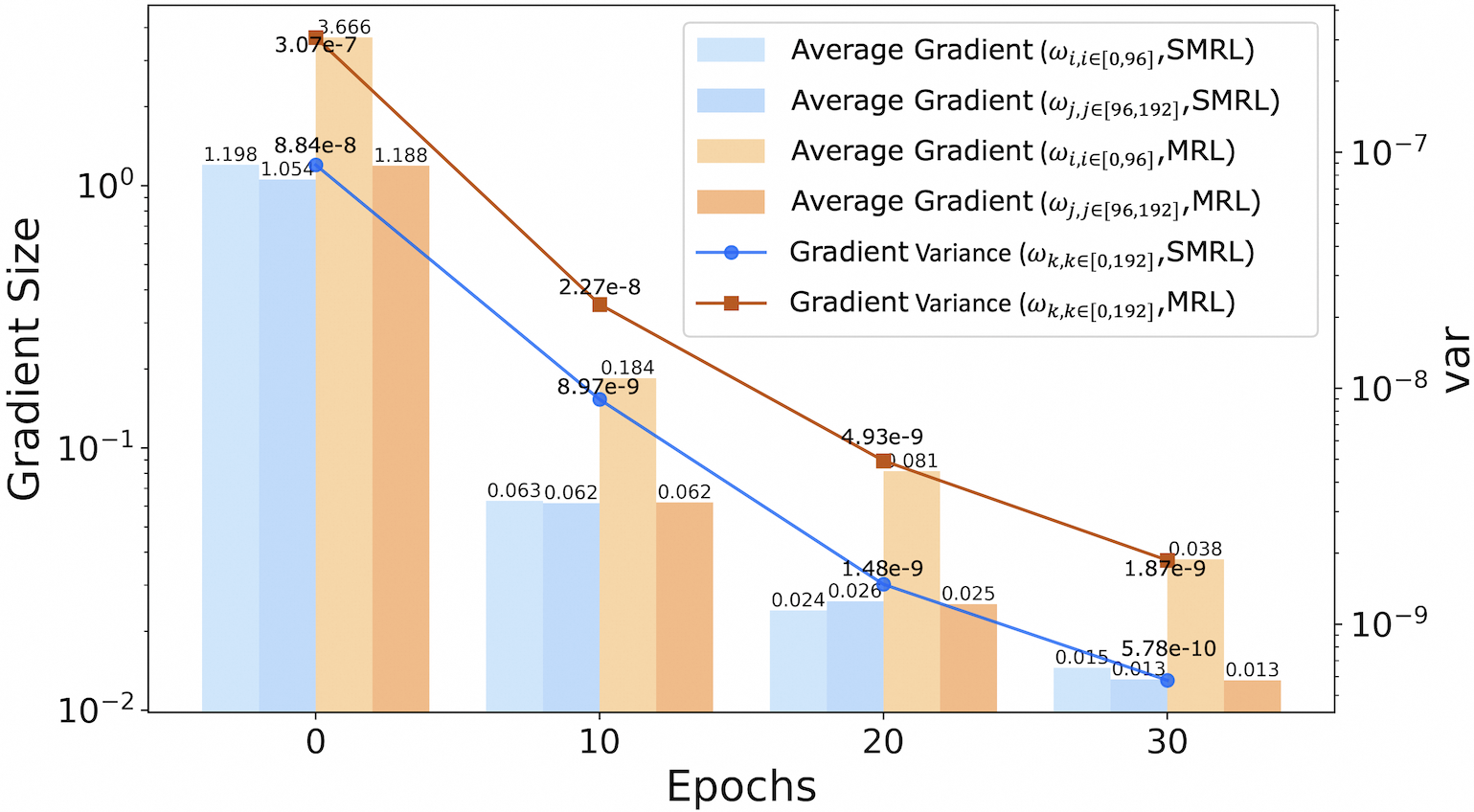}
    \caption{Rank Loss}
    \label{fig:rankloss}
\end{subfigure}
\hfill
\begin{subfigure}[t]{0.48\textwidth}
    \centering
    \includegraphics[width=\linewidth]{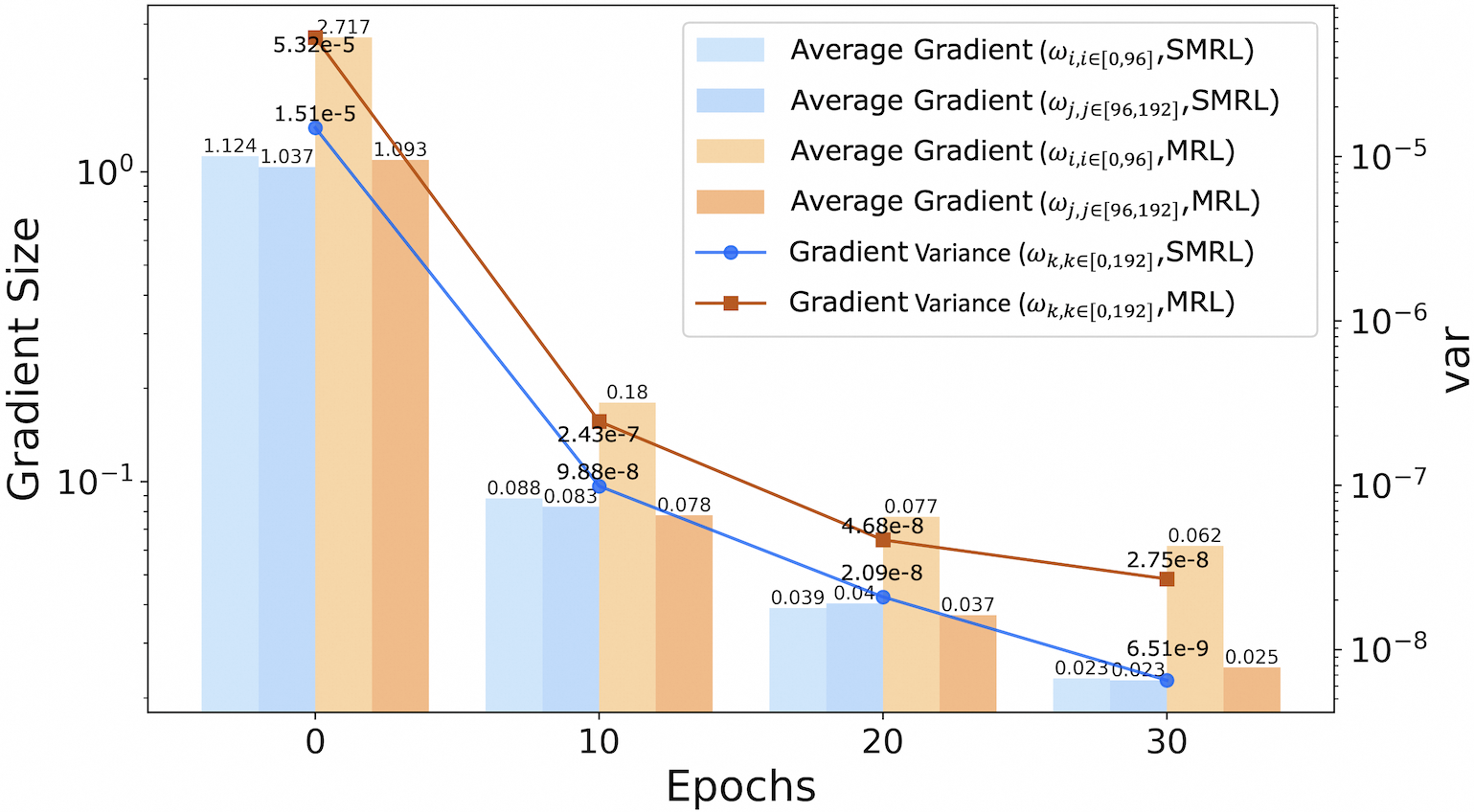}
    \caption{MSE Loss}
    \label{fig:mseloss}
\end{subfigure}
\hfill
\begin{subfigure}[t]{0.48\textwidth}
    \centering
    \includegraphics[width=\linewidth]{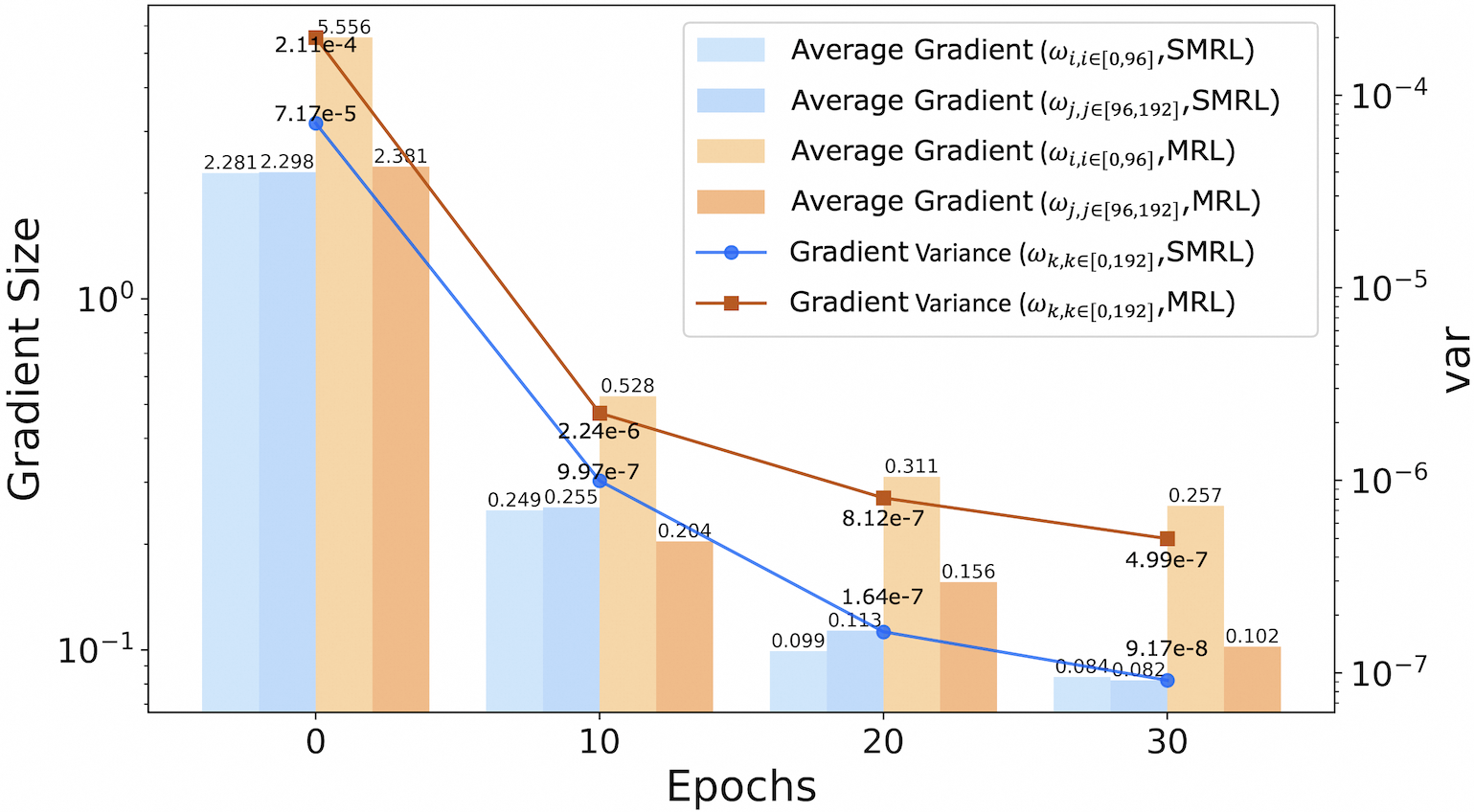}
    \caption{CE Loss}
    \label{fig:celoss}
\end{subfigure}
\caption{Gradient statistics with Rank, MSE and CE loss (with the vertical axis in a logarithmic scale): average gradient magnitudes of parameters in the ranges $[0, 96]$ and $[96, 192]$, as well as the gradient variance over all parameters in the range $[0, 192]$, during training.
}
\label{fig:discuss2}
\end{figure}

\subsection{Gradient variance of different loss functions}
\label{sec5.2}
Section \ref{sec5.1} demonstrates that MRL exhibits higher gradient variance compared to SMRL when rank loss is employed as the loss function, thereby corroborating the findings presented in Section \ref{sec3.2}. To enhance the validation, we conducted additional experiments on the BEIR dataset using rank loss, MSE loss and cross-entropy (CE) loss under identical settings. The results depicted in Figure \ref{fig:discuss2} reveal a consistent pattern across both loss functions, validating the robustness of our conclusions.

\subsection{Ablation studies} To evaluate the contribution of each component in SMEC to the overall performance, we conduct ablation studies using MRL as the baseline. Different modules are incrementally added on top of MRL, as detailed in table \ref{tab1}. When examined individually, the SMRL strategy achieves the most significant performance gain, suggesting that its reduced gradient variance contributes positively to model performance. In addition, both the ADS module and the S-XBM module also provide notable improvements. The combination of all three components improves the performance of the 128-dimensional embedding by 3.1 points.

% To evaluate the impact of gradient variance on model convergence speed and performance (see Section \ref{sec3.2} for details), we conducted comparative experiments between SMRL and MRL (with a target dimension of 96) based on the MiniLM model on the BEIR dataset. As shown in Figure~\ref{fig:figure6}(a), the gradient variance of the first 192 parameters during training is presented, and notably, the gradient variance of MRL remains several times higher than that of SMRL throughout the training process. In contrast, while the training loss of MRL continues to decrease after the 20th epoch, SMRL reaches a stable and fluctuating convergence state around the 15th epoch with a lower loss value. A similar trend can be observed in Figure~\ref{fig:figure6}(c), where SMRL begins to improve its performance earlier and achieves better final performance after convergence.

\begin{table}[h]
\centering
\resizebox{0.48\textwidth}{!}{%
\begin{tabular}{lcccc}
\hline
Method & 64 & 128 & 256 & 512 \\
\hline
MRL (Baseline) & 0.3726 & 0.4534 & 0.4802 & 0.5207 \\
\hline
w/ SMRL       & 0.3808 & 0.4621 & 0.4895 & 0.5283 \\
w/ ADS        & 0.3765 & 0.4583 & 0.4863 & 0.5254 \\
w/ S-XBM      & 0.3778 & 0.4583 & 0.4853 & 0.5256 \\
\rowcolor[gray]{0.9} SMEC (Ours) & \textbf{0.4053} & \textbf{0.4848} & \textbf{0.5002} & \textbf{0.5459} \\
\hline
\end{tabular}%
}
\caption{Ablation studies of SMEC on 8 BEIR datasets with MRL as the baseline.}
\label{tab1}
\end{table}

\subsection{The contribution of ADS in preserving key information}

The selection of important parameters in neural networks is a well-established research area, with numerous studies demonstrating that network parameters are often redundant. As a result, Parameter Pruning have been widely adopted for model compression. We consider ADS (or more generally, the MEC family of methods), although it focuses on dimension selection within embeddings, to be fundamentally implemented through network parameter selection. Therefore, ADS can be regarded as a form of pruning method with theoretical feasibility.

To fully demonstrate the effectiveness of ADS, we evaluate both the dimension selection strategies of ADS and MRL using WARE\cite{yu2018nisp} (Weighted Average Reconstruction Error), a commonly used metric in the pruning area for assessing parameter importance. The WARE is defined as follows:

\begin{equation}
\label{eq14}
\text{WARE} = \frac{1}{M} \sum_{m=1}^{M} \frac{|\hat{y}_m - y_m|}{|y_m|}
\end{equation} ,where $M$ denotes the number of samples; $\hat{y}_m$ and $y_m$ represent the model's score (which can be interpreted as the similarity between embedding pairs) for the $m$-th sample before and after dimension pruning, respectively. The core idea of WARE is to quantify the change in the model's output induced by removing a specific dimension; a larger change indicates higher importance of that dimension.

We randomly sampled 10,000 instances from multiple sub-datasets of BEIR. For the LLM2VEC embeddings (3072dim), we computed the WARE for each dimension. Then, we used both ADS and MRL to generate low-dimensional embeddings of 1536, 768, and 256 dimensions, respectively. For each method and compression level, we calculated the achievement rate, which is defined as the proportion of selected dimensions that appear in the top-N most important dimensions according to the WARE-based ranking.

\begin{table}[h]
\centering
\resizebox{0.48\textwidth}{!}{%
\begin{tabular}{lcc}
\toprule
\textbf{Dimension} & \textbf{ADS (Dimension Selection)} & \textbf{MRL (Dimension Truncation)} \\
\midrule
1536 & 94.3\% & 50.3\% \\
768 & 90.1\% & 32.8\% \\
256 & 83.6\% & 17.4\% \\
\bottomrule
\end{tabular}
}
\caption{Achievement Rate of Important Dimension Selection at Different Dimension Levels.} 
\label{tab3}
\end{table}

The results in table \ref{tab3} show that the achievement rate of MRL is roughly linear with the compression ratio, indicating that the importance of dimensions has no strong correlation with their positions. The achievement rate of ADS also decreases as the number of retained dimensions reduces, which is due to the increased difficulty of selecting the top-N most important dimensions under higher compression ratios. However, even when compressed by a factor of 6, ADS still selects over 80  of the most important dimensions. This explains why, as seen in Figure \ref{fig:exp2_1}, SMEC demonstrates stronger performance at lower dimensions.

\subsection{Memory size of S-XBM} In this subsection, we explore how the memory size of S-XBM module affects training speed and model performance. Theoretically,  as the memory size increases, it is easier for the S-XBM module to mine more hard samples, thereby improving model performance. However, an excessively large memory size may increase the retrieval time for top-k samples, which could negatively affect training efficiency. To prove this observation experimentally,  we train the SMEC framework with varying memory sizes (e.g., 1000, 2000, 5000, 10000, and 15000), as illustrated in the table \ref{tab2}. The results demonstrate a clear trade-off between training speed and model performance. We select a memory size of 5000 as our final choice to strike a balance between them.

\begin{table}[h]
\centering
\resizebox{0.48\textwidth}{!}{%
\begin{tabular}{lccccc}
\hline
Memory Size & 1000 & 2000 & 5000 & 10000 & 15000 \\  
\hline
Forward Time/s $\downarrow$ & 0.06 & 0.08 & 0.11 & 0.15 & 0.21 
\\ 
NDCG@10 $\uparrow$ & 0.4631 & 0.4652 & 0.4675 & 0.4682 & 0.4689 
\\ \hline
\end{tabular}%
}
\caption{Trade-off analysis of training speed and model performance under different memory size of S-XBM.}
\label{tab2}
\end{table}

\section{Conclusions}
\label{sec:bibtex}
Although high-dimensional embeddings from large language models (LLMs) capture rich semantic features, their practical use is often limited by computational efficiency and storage constraints. To mitigate these limitations, Sequential Matryoshka Embedding Compression (SMEC) framework is proposed in this paper to achieve efficient embedding compression. Our proposed SMEC framework contains Sequential Matryoshka Representation Learning(SMRL) module, adaptive dimension selection (ADS) module and Selectable Cross-batch Memory (S-XBM) module. The SMRL module is designed to mitigate gradient variance during training. The ADS module is utilized to minimize information degradation during feature compression. And the S-XBM is utilized to enhance unsupervised learning between high- and low-dimensional embeddings. Compared to existing approaches, our approaches preserve higher performance at the same compression rate.

\section*{Limitations}

The SMEC framework introduces only a small number of additional parameters on top of a pre-trained model and is trained using labeled data from a specific domain, along with mined hard samples, with the aim of reducing the dimensionality of the original embeddings. However, this design and objective limit its generalizability and applicability to broader scenarios. Future work could explore extending the SMEC approach to full-parameter training of representation models, enabling them to directly generate embeddings of multiple dimensions. Additionally, the feasibility of training the model on diverse datasets is also worth investigating.

\bibliography{ref}

\onecolumn
\newpage

\appendix

\section{Derivation of the Gradient Fluctuation}
\label{appd:a}
To formalize this issue, we analyze the Mean Squared Error (MSE) loss as a representative case. Let \(\mathbf{x}_1 = [x_1, x_2, \ldots, x_n]^\top \in \mathbb{R}^n\) and \(\mathbf{x}_2 = [y_1, y_2, \ldots, y_n]^\top \in \mathbb{R}^n\) denote two input feature vectors. The final FC layer employs a matrix \(\mathbf{W} = [\mathbf{w}_1, \mathbf{w}_2, \ldots,\mathbf{w}_n]^\top \in \mathbb{R}^{m \times n}\) to generate scalar outputs \(\mathbf{y}_1 = \mathbf{W} \mathbf{x}_1 \in \mathbb{R}^{m} \) and \(\mathbf{y}_2 = \mathbf{W} \mathbf{x}_2 \in \mathbb{R}^{m} \). The MSE loss at dimension \(d\) is defined as:  
\begin{equation}
\label{eq1}
\mathcal{L}^{d} = \left[
\mathcal{Y}_{label} - sim( \mathbf{y}^{d}_1,\mathbf{y}^{d}_2)
\right]^2 ,
\end{equation} where \( \mathcal{Y}_{label} \) denotes the binary classification label for pairs (0 or 1), and $sim(\cdot)$ represents the normalized similarity of the learned representations. 

According to the chain rule, the partial derivative of \(\mathcal{L}^d\) with respect to the \(i\)-th dimension parameter of the FC layer is derived as:
\begin{equation}
\label{eq2}
\frac{\partial \mathcal{L}^{d}}{\partial \mathbf{w}_i} = \frac{\partial \mathcal{L}^{d}}{\partial \left[\mathbf{y}^{d}_1\right]_i} \cdot \frac{\partial \left[\mathbf{y}^{d}_1\right]_i}{\partial \mathbf{w}_i} + \frac{\partial \mathcal{L}^{d}}{\partial \left[\mathbf{y}^{d}_2\right]_i} \cdot \frac{\partial \left[\mathbf{y}^{d}_2\right]_i}{\partial \mathbf{w}_i} .
\end{equation}

Utilizing cosine similarity (clamp to $[0,1]$) as the similarity function \( sim(\cdot) \), the equation \ref{eq1} can be rewritten as:
 
\begin{equation}
\label{eq3}
\mathcal{L}^{d} = \left[
\mathcal{Y}_{label} - \frac{{\mathbf{y}^{d}_1}^\top {\mathbf{y}^{d}_2}}
{\| \mathbf{y}^{d}_1 \| \| \mathbf{y}^{d}_2 \|}
\right]^2 .
\end{equation}

Let $\|\mathbf{y}^d_1\| = A$, $\|\mathbf{y}^d_2\| = B$ , ${\mathbf{y}^{d}_1}^\top {\mathbf{y}^{d}_2} = C$ and $s = \frac{C}{AB}$. The partial derivatives of the $\mathcal{L}^{d}$ with respect to $\left[\mathbf{y}^{d}_1\right]_i$ and $\left[\mathbf{y}^{d}_2\right]_i$ are given as follows:

\begin{equation}
\label{eq4}
\frac{\partial \mathcal{L}^{d}}{\partial \left[\mathbf{y}^{d}_1\right]_i} 
= 2\left( s-\mathcal{Y}_{label}\right) \left( \frac{ \left[\mathbf{y}^{d}_2\right]_i }{AB} - \frac{s}{A^2} \left[\mathbf{y}^{d}_1\right]_i \right) ,
\end{equation}

\begin{equation}
\label{eq5}
\frac{\partial \mathcal{L}^{d}}{\partial \left[\mathbf{y}^{d}_2\right]_i} = 2\left( s-\mathcal{Y}_{label}\right) \left( \frac{\left[\mathbf{y}^{d}_1\right]_i}{AB} - \frac{s}{B^2} \left[\mathbf{y}^{d}_2\right]_i \right) .
\end{equation}

Substituting $\left[\mathbf{y}^{d}_1\right]_i = \mathbf{w}_i\mathbf{x}_1$ and $\left[\mathbf{y}^{d}_2\right]_i = \mathbf{w}_i\mathbf{x}_2$, the partial derivatives of the $ \left[\mathbf{y}^{d}_1\right]_i$ and $\left[\mathbf{y}^{d}_2\right]_i$ with respect to $\mathbf{w}_i$ are given as follows:

\begin{equation}
\label{eq6}
\frac{\partial \left[\mathbf{y}^{d}_1\right]_i}{\partial \mathbf{w}_i} = \mathbf{x}_1 , \frac{\partial \left[\mathbf{y}^{d}_2\right]_i}{\partial \mathbf{w}_i} = \mathbf{x}_2 .
\end{equation}

Based on the above equations, the partial derivative of $\mathcal{L}^d$ with respect to $\mathbf{w}_i$ is derived as:

\begin{align}
\label{eq7}
\frac{\partial \mathcal{L}^{d}}{\partial \mathbf{w}_i} 
=& \ 2\left( s-\mathcal{Y}_{label}\right) \left[ \left( \frac{ \left[\mathbf{y}^{d}_2\right]_i }{AB} - \frac{s}{A^2} \left[\mathbf{y}^{d}_1\right]_i \right)\mathbf{x}_1 +  \left( \frac{\left[\mathbf{y}^{d}_1\right]_i}{AB} - \frac{s}{B^2} \left[\mathbf{y}^{d}_2\right]_i \right)\mathbf{x}_2 . \right]
\end{align}

Assume that $A$ and $B$ can be approximated by \( \delta(d) \cdot a \) and \( \delta(d) \cdot b \), respectively. Under this approximation, \( \delta(d) \) can be used to fit the relationship between the magnitude of vector \( \mathbf{x} \) or \( \mathbf{y} \) and the variation of \( d \) (It is evident that this is a positive correlation). Therefore, Equation \ref{eq7} can be approximated by the following expression:

\begin{equation}
\frac{\partial \mathcal{L}^{d}}{\partial \mathbf{w}_i}
= 2\left( s-\mathcal{Y}_{label}\right) \frac{1}{\delta(d)^2} \Bigg[
\left( \frac{ \left[\mathbf{y}^{d}_2\right]_i }{ab} - \frac{s}{a^2} \left[\mathbf{y}^{d}_1\right]_i \right)\mathbf{x}_1 + \left( \frac{\left[\mathbf{y}^{d}_1\right]_i}{ab} - \frac{s}{b^2} \left[\mathbf{y}^{d}_2\right]_i \right)\mathbf{x}_2
\Bigg].
\label{eq8}
\end{equation}

In equation \ref{eq8}, $a$, $b$, $\left[\mathbf{y}^{d}_1\right]_i $, $\left[\mathbf{y}^{d}_2\right]_i $ are constants, $\mathbf{x}_1$ and $\mathbf{x}_2$ are constant vectors, while $s$ and $\mathcal{Y}_{label}$ are invariant with respect to the index $d$. Therefore, we can conclude the following:

\begin{equation}
\label{eq9}
\frac{\partial \mathcal{L}^{d}}{\partial \mathbf{w}_i} \propto \frac{1}{\delta(d)^2}.
\end{equation} In theory, this rule can also be extended to other pair-wise similarity-based functions, such as rank loss, which is experimentally verified in Section \ref{sec5.2}.

% This formulation provides a mathematical foundation for analyzing gradient fluctuation in shared-parameter architectures. It demonstrates that in the MRL architecture, loss functions across different dimensions result in gradients of varying magnitudes for the same model parameter, thereby increasing gradient variance. In Section \ref{sec3.2}, we propose a solution to address this issue.

\section{Results on BEIR Sub-datasets.}

We compare the performance of different compression methods on several representative sub-datasets of BEIR, and the results are shown in Table \ref{beir-sub}.

\begin{table}[h]
\centering
\begin{tabular}{lcccccc}
\toprule
\textbf{} & \multicolumn{6}{c}{\textbf{NDCG@10}} \\

\cmidrule(lr){2-7}

\textbf{Model} & \textbf{128} & \textbf{256} & \textbf{512} & \textbf{768} & \textbf{1536} & \textbf{3072} \\
\hline
\multicolumn{7}{l}{\textit{Sub-dataset--Scifact}}  \\
LLM2Vec & - & - & - & - & - & 0.787 \\
w/ Search-Adaptor & 0.806 & 0.845 & 0.864 & 0.879 & 0.886 & 0.884 \\
w/ MRL-Adaptor & 0.826 & 0.861 & 0.876 & 0.880 & 0.886 & 0.887 \\
w/ SMEC (ours) & 0.841 & 0.874 & 0.879 & 0.882 & 0.885 & 0.886 \\
\midrule
\multicolumn{7}{l}{\textit{Sub-dataset--FiQA}}  \\
LLM2Vec & - & - & - & - & -  & 0.498 \\
w/ Search-Adaptor & 0.475 & 0.505 & 0.529 & 0.540 & 0.545 & 0.550 \\
w/ MRL-Adaptor & 0.496 & 0.523 & 0.534 & 0.543 & 0.547 & 0.550 \\
w/ SMEC (ours) & 0.521 & 0.533 & 0.540 & 0.546 & 0.549 & 0.551 \\
\bottomrule
\multicolumn{7}{l}{\textit{Sub-dataset--Quora}}  \\
LLM2Vec & - & - & - & - & -  & 0.775 \\
w/ Search-Adaptor & 0.771 & 0.805 & 0.830 & 0.845 & 0.861 & 0.864 \\
w/ MRL-Adaptor & 0.784 & 0.812 & 0.834 & 0.847 & 0.862 & 0.863 \\
w/ SMEC (ours) & 0.794 & 0.818 & 0.839 & 0.850 & 0.862 & 0.865 \\
\bottomrule
\multicolumn{7}{l}{\textit{Sub-dataset--NFCorpus}}  \\
LLM2Vec & - & - & - & - & -  & 0.389 \\
w/ Search-Adaptor & 0.345 & 0.375 & 0.396 & 0.412 & 0.425 & 0.426 \\
w/ MRL-Adaptor & 0.364 & 0.384 & 0.403 & 0.419 & 0.426 & 0.427 \\
w/ SMEC (ours) & 0.389 & 0.402 & 0.418 & 0.426 & 0.430 & 0.431 \\

\bottomrule
\multicolumn{7}{l}{\textit{Sub-dataset--SciDocs}}  \\
LLM2Vec & - & - & - & - & -  & 0.232 \\
w/ Search-Adaptor & 0.204 & 0.225 & 0.245 & 0.250 & 0.258 & 0.263 \\
w/ MRL-Adaptor & 0.220 & 0.240 & 0.250 & 0.255 & 0.262 & 0.265 \\
w/ SMEC (ours) & 0.239 & 0.246 & 0.251 & 0.255 & 0.261 & 0.264 \\
\bottomrule
\end{tabular}
\caption{Comparison of retrieval performance on 5 BEIR sub-datasets.}
\label{beir-sub}
\end{table}

\section{Experimental results on Products-10K and Fashion-200k.}
\label{appd:b}
\begin{figure}
\centering
\begin{subfigure}[t]{0.32\textwidth}
    \centering
    \includegraphics[width=\linewidth]{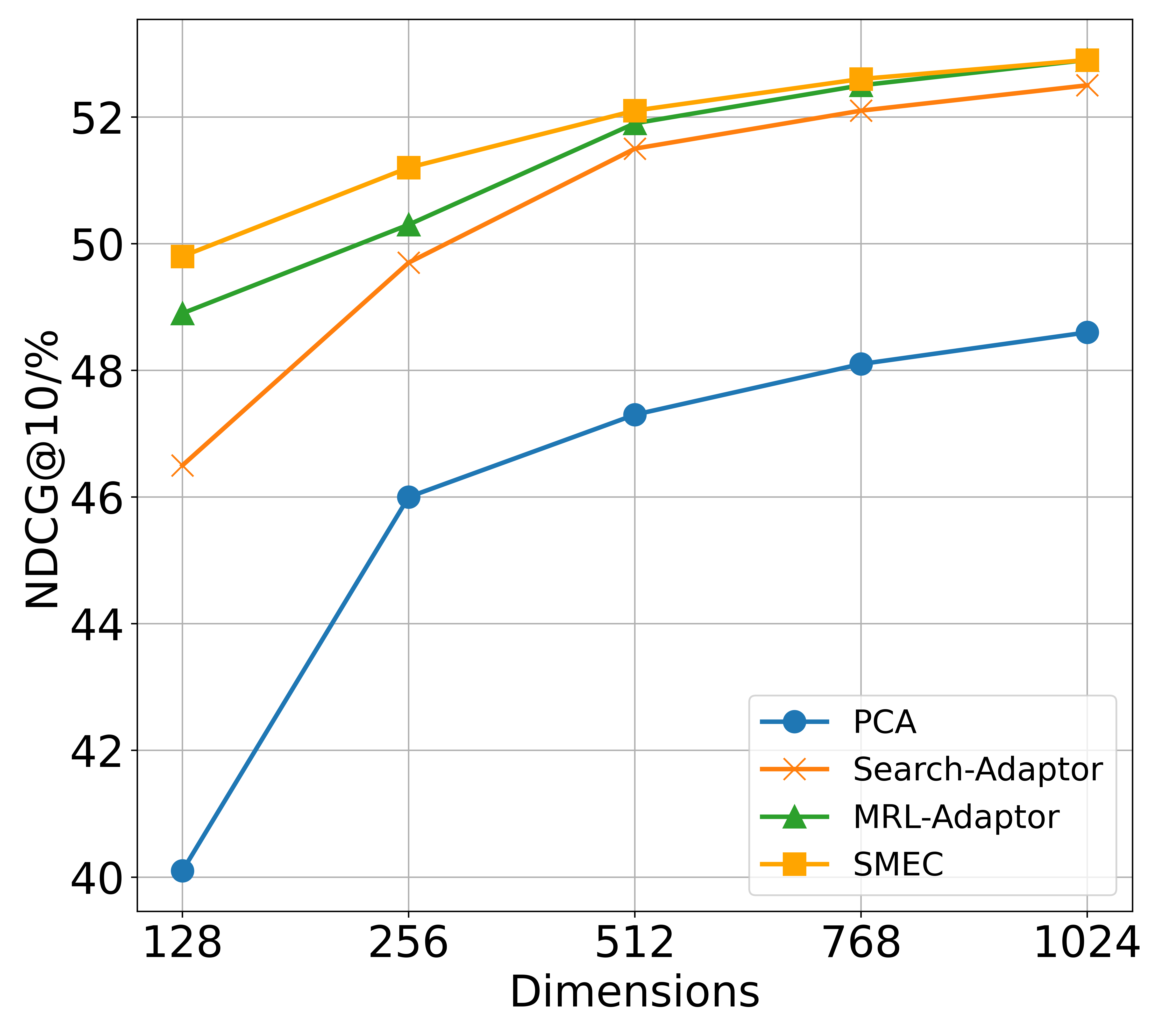}
    \caption{Image retrieval}
    \label{fig:img}
\end{subfigure}
\hfill
\begin{subfigure}[t]{0.32\textwidth}
    \centering
    \includegraphics[width=\linewidth]{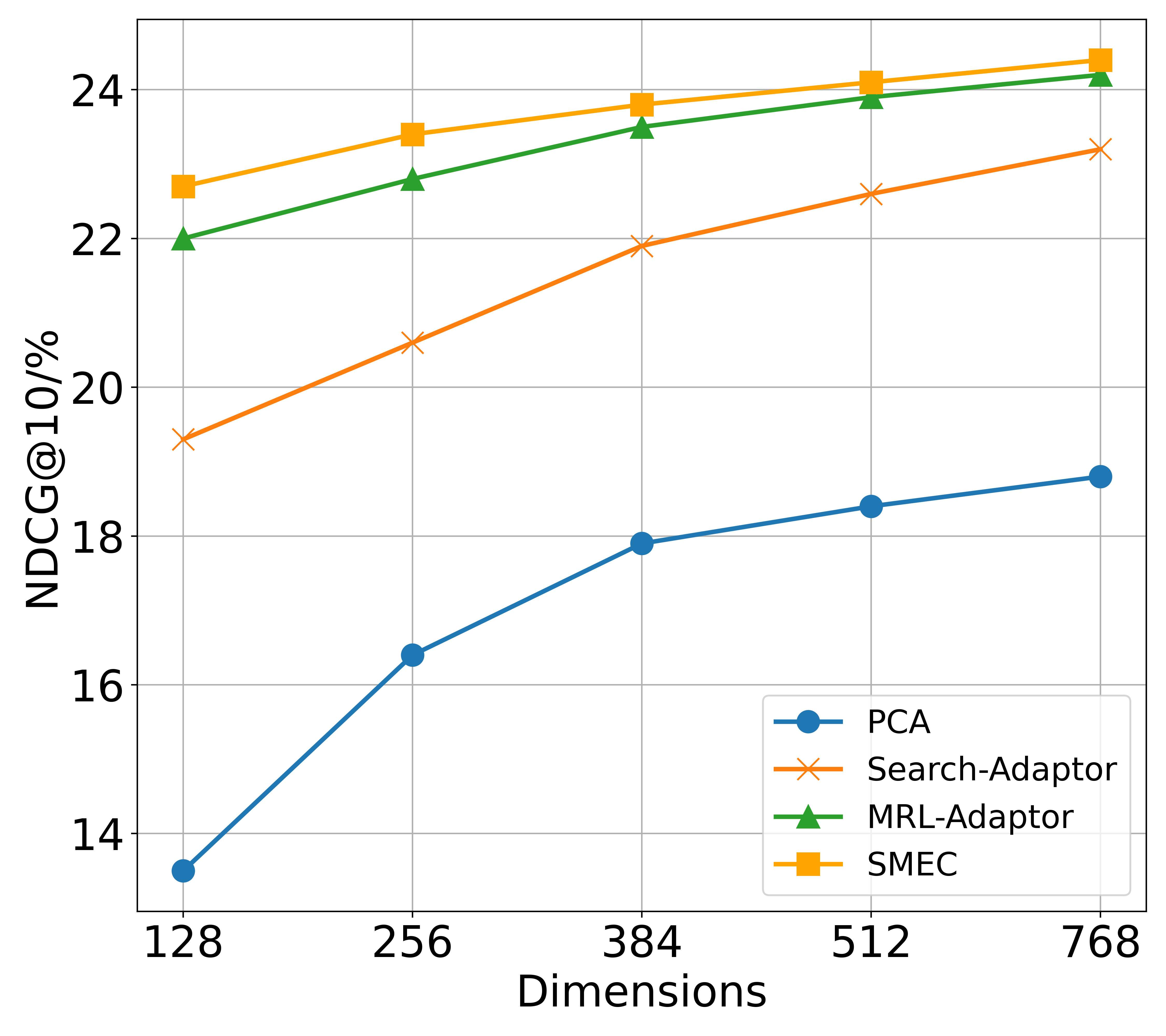}
    \caption{Text-to-Image retrieval}
    \label{fig:t2i}
\end{subfigure}
\hfill
\begin{subfigure}[t]{0.32\textwidth}
    \centering
    \includegraphics[width=\linewidth]{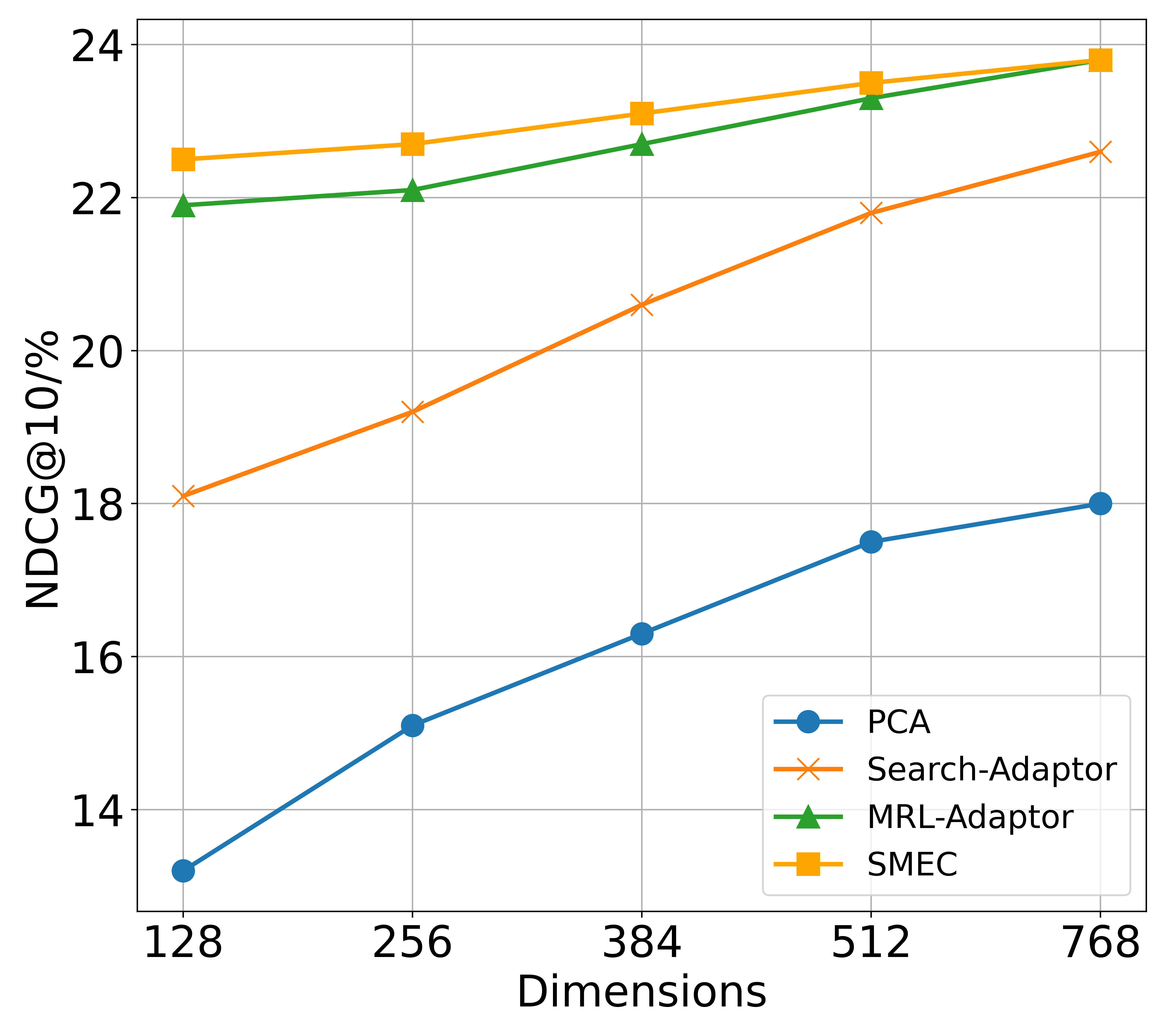}
    \caption{Image-to-Text retrieval}
    \label{fig:i2t}
\end{subfigure}
\caption{Experimental results on image and multimodal datasets. (a) presents the results on the Products-10K dataset using an image representation model based on ViT-H (with 1024 dimensions). (b) and (c) show the results on the Fashion-200K dataset for text-to-image and image-to-text retrieval tasks, respectively, using the LLM2CLIP model (with 768 dimensions, base on ViT-L/14 and Llama-3.2-1B).}
\label{fig:exp2_2}
\end{figure}

\end{document}